\documentclass{article}
\usepackage[top=1in, bottom=1in, left=1in, right=1in]{geometry} 
\usepackage{soul, color}
\usepackage{float}
\usepackage[utf8]{inputenc}
\usepackage{placeins}
\usepackage{booktabs}
\usepackage{makeidx}  % allows for indexgeneration
\usepackage[utf8]{inputenc}
\usepackage[misc]{ifsym}
\usepackage{soul}

\usepackage{graphicx}
\usepackage{comment}
\usepackage{amsmath,amssymb}
\usepackage{color}
\usepackage{tabularx}
\usepackage{multirow}
\usepackage{balance}
\usepackage{comment}   
\usepackage{amsfonts}
\usepackage{array}
\usepackage{url}
\usepackage{textcomp}
\usepackage{bm}

\usepackage{algorithm} 
\usepackage{algorithmic}
\usepackage{mathtools}
\DeclarePairedDelimiter\floor{\lfloor}{\rfloor}

\newcommand{\T}{\mathrm{T}}
\newcommand{\M}[1]{\mathbf{#1}}
\newcommand{\C}[1]{\mathcal{#1}}

\usepackage[tight]{subfigure}

\usepackage[pagebackref=true,breaklinks=true,colorlinks,bookmarks=false]{hyperref}
\usepackage[dvipsnames]{xcolor}
\hypersetup{
    colorlinks=true,       % false: boxed links; true: colored links
    linkcolor=red,          % color of internal links
    citecolor=NavyBlue,        % color of links to bibliography
}

\title{Adversarial Memory Networks for Action Prediction}
\author{Zhiqiang Tao\thanks{Equal contribution}~\thanks{Santa Clara University, email: \texttt{ztao@scu.edu}}
\and Yue Bai\footnotemark[1]~\thanks{Northeastern University, email: \texttt{bai.yue@northeastern.edu}}
\and Handong Zhao\thanks{Adobe Research, email: \texttt{hazhao@adobe.com}}
\and Sheng Li\thanks{University of Georgia, email: \texttt{sheng.li@uga.edu}}
\and Yu Kong\thanks{Rochester Institute of Technology, email: \texttt{yu.kong@rit.edu}}
\and Yun Fu\thanks{Northeastern University, email: \texttt{yunfu@ece.neu.edu}}
}

\date{}
\begin{document}
\maketitle

%%%%%%%%% ABSTRACT
\begin{abstract}
Action prediction aims to infer the forthcoming human action with partially-observed videos, which is a challenging task due to the limited information underlying early observations. Existing methods mainly adopt a reconstruction strategy to handle this task, expecting to learn a single mapping function from partial observations to full videos to facilitate the prediction process. In this study, we propose adversarial memory networks (AMemNet) to generate the ``full video'' feature conditioning on a partial video query from two new aspects. Firstly, a key-value structured memory generator is designed to memorize different partial videos as key memories and dynamically write full videos in value memories with gating mechanism and querying attention. Secondly, we develop a class-aware discriminator to guide the memory generator to deliver not only realistic but also discriminative full video features upon adversarial training. The final prediction result of AMemNet is given by late fusion over RGB and optical flow streams. Extensive experimental results on two benchmark video datasets, UCF-101 and HMDB51, are provided to demonstrate the effectiveness of the proposed AMemNet model over state-of-the-art methods.
\end{abstract}

%%%%%%%%% BODY TEXT
\section{Introduction}
Action prediction is a highly practical research topic that could be used in many real-world applications such as video surveillance, autonomous navigation, human-computer interaction, etc. Different from action recognition, which recognizes the human action category upon a complete video, action prediction aims to understand the human activity at an early stage -- using a partially-observed video before an entire action execution. Typically, action prediction methods~\cite{RyooICCV2011,CaoCVPR2013,KongCVPR2017,Chen_2018_ECCV,Wang_2019_CVPR,Zhao_2019_ICCV} assume that a portion of consecutive frames from the beginning is given, considered as a partial video. The challenges mainly arise from the limited information in the early progress of videos, leading to the incomplete temporal context and a lack of discriminative cues for recognizing actions. Thus, the key problem of solving action prediction lies in: \emph{how to enhance the discriminative information for a partial video}?

In recent years, many research efforts centering on the above question have been made for the action prediction task. Pioneering works~\cite{RyooICCV2011,CaoCVPR2013,KongECCV2014} mainly handle partial videos by relying on hand-crafted features, dictionary learning, and designing temporally-structured classifiers. More recently, deep convolutional neural networks (CNNs), especially those pre-trained on large-scale video benchmarks (\emph{e.g.}, Sports-1M~\cite{KarpathyCVPR14} and Kinetics~\cite{Kinetics}), have been widely adopted to predict actions. The pre-trained CNNs, to some extent, compensate for the incomplete temporal context and empower reconstructing full video representations from the partial ones. Along this line, existing methods~\cite{KongCVPR2017,Shi_2018_ECCV,Wang_2019_CVPR,AAPnet-TPAMI20} focus on designing models to continuously improve the reconstruction performance, yet without considering the ``malnutrition'' nature (\emph{i.e.}, the limited temporal cues) of incomplete videos. Particularly, it will be more straightforward to learn what ``nutrients'' (\emph{e.g.}, the missing temporal cues or reconstruction bases) a partial video may need for recognizing actions, compared with mapping it to an entire video. Moreover, it is also challenging to handle various partial videos by resorting to a single model.

In this study, we propose a novel adversarial memory networks (AMemNet) model to address the above challenges. The proposed AMemNet leverages augmented memory networks to explicitly learn and store full video features to enrich incomplete ones. Specifically, we treat a partial video as a \emph{query} and the corresponding full video as its \emph{memory}. The ``full video'' is generated with relevant memory slots fetched by the query of partial videos. We summarize the contribution of this work in two aspects.

Firstly, a memory-augmented generator model is designed for generating full-video features conditioning on partial-video queries. We adopt a key-value memory network architecture~\cite{KvMemNet,KVMemNet-WWW17} for action prediction, where the key memory slots are used for capturing similar partial videos and the value memory slots are extracted from full training videos. The memory writing process is implemented by gating mechanism and attention weights. The input/forget gates enable AMemNet to dynamically update video memories attended by different queries and thus memorize the variation between different video progress.

Secondly, a class-aware discriminator model is developed to guide the memory generator with adversarial training, which not only employs an adversarial loss to encourage generating realistic full video features, but also imposes a classification loss on training the network. By this means, the discriminator network could further push the generator to deliver discriminative full-video features.

The proposed AMemNet obtains prediction results by employing a late fusion strategy over two streams (\emph{i.e.}, RGB and optical flow) following~\cite{SimonyanNIPS2014,WangECCV2016}. Extensive experiments on two benchmark datasets, UCF101 and HMDB51, are conducted to show the effectiveness of AMemNet compared with state-of-the-art methods, where our approach surprisingly achieves over $90\%$ accuracy by only observing $10\%$ of the beginning video frames on the UCF101 dataset. A detailed ablation study compared with several competitive baselines is also presented.

\section{Related Work}
\textbf{Action Recognition} targets at recognizing the label of human action in a given video, which is one of the core tasks for video understanding. Previous works have extensively studied this research problem from several aspects, including hand-crafted features (\emph{e.g.}, spatio-temporal interest points~\cite{Dollar2005,RyooICCV2009}, poselet key-frames~\cite{RaptisCVPR2013,LaptevIJCV2005}), and dense trajectory~\cite{WangIJCV2013}, 3D convolutional neural networks~\cite{JiPAMI2013,TranICCV2015,Hara_2018_CVPR}, recurrent neural networks (RNN) based methods~\cite{NgCVPR2015,Donahue_2015_CVPR}, and many recent deep CNN based methods such as temporal linear encoding networks~\cite{Diba_2017_CVPR}, non-local neural networks~\cite{Wang_2018_CVPR}, etc. Among existing methods, the two-stream architecture~\cite{SimonyanNIPS2014,Feichtenhofer_2016_CVPR,WangECCV2016} forms a landmark~\cite{CarreiraCVPR2017}, which mainly employ deep CNNs on the RGB and optical flow streams for exploiting the spatial-temporal information inside videos. In this work, we also adopt the two-stream structure as it naturally provides the complimentary information for the action prediction task -- the RGB stream contributes more on the early observation and the optical flow leads the following progress.

\textbf{Action Prediction} has attracted lots of research efforts~\cite{KongCVPR2017,Shi_2018_ECCV,Wang_2019_CVPR,AAPnet-TPAMI20,Chen_2018_ECCV,Zhao_2019_ICCV} in recent years, which tries to predict action labels upon the early progress of videos and thus falls into a special case of video-based action recognition. Previous works~\cite{RyooICCV2011,CaoCVPR2013,LanECCV2014,KongECCV2014} solve this task via hand-crafted features, and recent works~\cite{KongCVPR2017,Shi_2018_ECCV,Wang_2019_CVPR,AAPnet-TPAMI20,TSL-wacv18,KongAAAI2018,GuoQMD18,Chen_2018_ECCV,Zhao_2019_ICCV} mainly rely on pre-trained deep CNN models for encoding videos, such as the 3D convolutional networks in~\cite{KongCVPR2017,AAPnet-TPAMI20,Wang_2019_CVPR}, deep CNNs in~\cite{Shi_2018_ECCV,Chen_2018_ECCV}, and two-stream CNNs in~\cite{TSL-wacv18,KongAAAI2018,GuoQMD18,Zhao_2019_ICCV}. Among these works, the most common way for predicting actions is to design deep neural networks model for reconstructing full videos from the partial ones, such as deep sequential context networks~\cite{KongCVPR2017}, the RBF Kernelized RNN~\cite{Shi_2018_ECCV}, progressive teacher-student learning networks~\cite{Wang_2019_CVPR}, adversarial action prediction networks~\cite{AAPnet-TPAMI20}, etc. Moreover, some other interesting methods include the LSTM based ones~\cite{Aliakbarian_2017_ICCV,KongAAAI2018,wang2019eidetic}, part-activated deep reinforcement learning~\cite{Chen_2018_ECCV}, residual learning~\cite{GuoQMD18,Zhao_2019_ICCV}, motion prediction~\cite{ECCV20-ref1}, asynchronous activity anticipation~\cite{ECCV20-ref2}, etc.

The memory augmented LSTM (Mem-LSTM)~\cite{KongAAAI2018} model and adversarial action prediction networks (AAPNet)~\cite{AAPnet-TPAMI20} share some similar ideas to our approach. However, several essential differences between the proposed AMemNet and Mem-LSTM/AAPnet could be summarized. First, memory networks play distinct roles in Mem-LSTM~\cite{KongAAAI2018} and AMemNet. 
Mem-LSTM formulates action labels as video memories and adopts the memory network as a nearest-neighbor classifier. Differently, the proposed AMemNet develops a key-value memory architecture as a generator model and learns value memory slots from full videos as reconstruction bases for a generation purpose. Second, the generator models used in AAPNet~\cite{AAPnet-TPAMI20} and AMemNet are different. AAPNet~\cite{AAPnet-TPAMI20} employs a variational-GAN model, whereas AMemNet develops the memory-augmented generator to explicitly provide auxiliary information to generate full-video features for testing videos. 

\textbf{Memory Networks}, \emph{i.e.}, Memory-Augmented Neural Networks~\cite{WestonCB14,NTM-GravesWD14}, generally consist of two components: 1) a memory matrix and 2) a neural network controller, where the memory matrix is used for storing the information as memory slots and the neural network controller is generally designed for addressing, reading and writing memories. Several representative memory network architectures include end-to-end memory networks~\cite{end2endMemNet}, Key-Value memory networks~\cite{KvMemNet,KVMemNet-WWW17}, neural tuning machines~\cite{NTM-GravesWD14,WangYHLWH18}, recurrent memory networks~\cite{recMemNet,Tao0WFYZ019}, etc. The memory networks work well in practice for its flexibility in saving the auxiliary knowledge and its ability in memorizing the long-term temporal information. The proposed AMemNet shares the same memory architecture with~\cite{KvMemNet,KVMemNet-WWW17} and employs the memory writing methods provided in~\cite{NTM-GravesWD14}, which, however, is designed with different purposes compared with these methods. The memory module in AMemNet is tailored for solving the action prediction problem.

\begin{figure*}[t]
\centering
\includegraphics[width=0.96\textwidth]{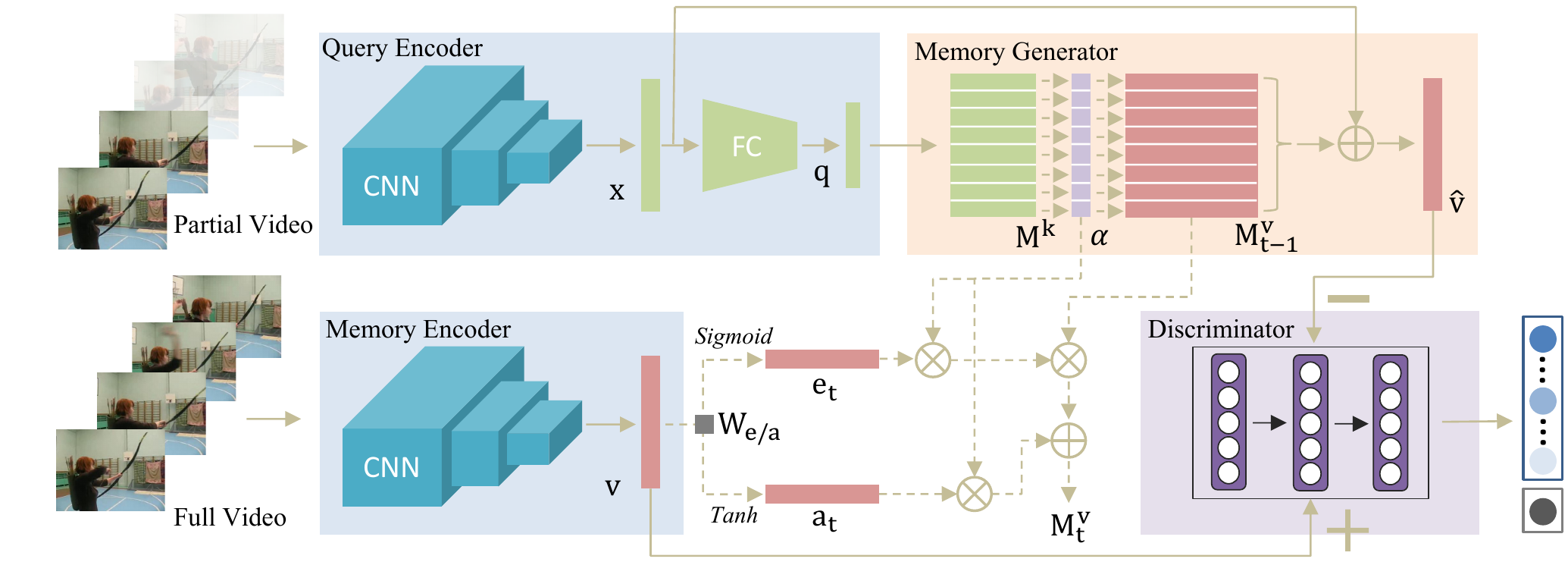}
\caption{Illustration of training the proposed AMemNet on the RGB stream. The attention weight $\alpha$ is first given by a query embedding $\M{q}$ and a key memory matrix $\M{M}^{k}$ in the memory addressing. Then, $\alpha$ is used for updating the value memory $\M{M}^{v}_{t-1} \rightarrow \M{M}^{v}_{t}$ with real full video features $\M{v}$ governed by input/forget gates $\M{W}_{e/a}$. The generated full video feature $\M{\hat{v}}$ is obtained by a memory reading operation with $\alpha$ and $\M{M}^{v}_{t}$, and guided by a class-aware discriminator network $D=\{D_{cls}, D_{adv}\}$ under adversarial training. The memory update (dash lines) will be disabled in testing.}\label{fig:framwork}
\end{figure*}

\section{Methodology}
\subsection{Problem Setup}\label{sec:setting}
Given a set of training videos $\{(x,y)\}$, where $x \in \C{X}$ denotes one video sample and $y \in \C{Y}$ refers to its action category label, action prediction aims to infer $y$ by only observing the beginning sequential frames of $x$ instead of using the entire video. Let $\tau \in (0,1]$ be the observation ratio and $L$ be the length (i.e., the total number of frames) of $x$, a \emph{partial video} is defined as $x_{1:\floor*{\tau L}}$ that is a subsequence of the full video $x$ containing from the first frame to the $\floor*{\tau L}$-th frame. We employ a set of observation ratios $\{\tau_p\}_{1}^{P}$ to mimic the partial observations at different progress levels and define $x_p=x_{1:\floor*{\tau_p L}}$ as the $p$-th progress level observation of $x$, $p \in \{1,\dots, P\}$. By this means, the training set is augmented as $P$ times of the original one, \emph{i.e.}, $\{(x_p,y)\}$. Following the existing work~\cite{KongCVPR2017,Chen_2018_ECCV,Zhao_2019_ICCV}, we set $P=10$ and increase $\tau_p$ from $0.1$ to $1.0$ with a fixed step of $0.1$. 

We propose to solve the action prediction problem with memory networks. The partial video $x_p$ is formulated as a \emph{query} to ``retrieve'' its lost information from a \emph{memory} block learned from all the full training videos. To encode video memories, we build a training set as $\{(x_p, v, y)\}$, where $v$ indicates the full video of $x_p$, \emph{i.e.}, $v:= x_P$. Different from previous works~\cite{KongCVPR2017,KongAAAI2018,AAPnet-TPAMI20}, which require the progress level $p$ during the training process, the proposed AMemNet is ``progress-free''. Hence, for convenience, we omit the subscript $p$ of the partial video $x_p$ when no confusion occurs, and always denote $(x, v, y)$ as a triplet of partial observation, full observation, and action category of the same video sample throughout the paper.

Following~\cite{SimonyanNIPS2014,WangECCV2016}, we train the proposed AMemNet model on the RGB frames and optical flows, respectively, and employ a late fusion mechanism to exploit the spatial-temporal information in a two-stream framework. Each stream employs the same network architecture with its own trainable weights. We refer to $(x^{rgb}, v^{rgb}, y)$ / $(x^{flow}, v^{flow}, y)$ as two modalities, and omit the subscripts ($rgb$/$flow$) when it is unnecessary.

\subsection{Adversarial Memory Networks (AMemNet)}
Fig.~\ref{fig:framwork} shows the network architecture of the proposed AMemNet model. Overall, our model consists of three components: 1) query/memory encoder; 2) memory generator, and 3) discriminator. The encoder network is used to vectorize the given partial/full video as feature representations, the memory generator is learned to generate a full video representation conditioning on the partial video query, and the discriminator is trained to distinguish between \emph{fake} and \emph{real} full video representations and also deliver the prediction scores over all the categories. During the training process, the value memories are continuously updated by full videos with erase/add vectors that work as input/forget gates. We show the details of each component in the following.

\subsubsection{Query/Memory Encoder}
Given a partial video $x$ and its corresponding full video $v$, we employ deep convolution neural networks (CNN) as an encoding model to obtain feature representations as follows:
$\M{x} = f_{cnn}(x; \theta_{cnn})$ and $\M{v} = f_{cnn}(v; \theta_{cnn})$, where $\M{x} \in \mathbb{R}^{d}$ and $\M{v} \in \mathbb{R}^{d}$ are the encoded representations for $x$ and $v$, respectively, $d$ is the feature dimension, and $\theta_{cnn}$ parameterizes the CNN model. Following~\cite{TSL-wacv18,Zhao_2019_ICCV}, we instantiate $f_{cnn}(\cdot;\theta_{cnn})$ with the pre-trained TSN model~\cite{WangECCV2016} for its robust and competitive performance on action recognition. 

The proposed AMemNet model utilizes the partial video feature $\M{x}$ as a query to fetch relevant memories, which are learned from full training videos, to generate its full video feature $\M{v}$. Hence, it is natural to directly utilize $f_{cnn}(\cdot;\theta_{cnn})$ as the memory encoder for learning memory representations of full videos . On the other hand, to facilitate the querying process, we further encode the partial video representation $\M{x}$ in a lower-dimensional embedding space by
\begin{equation}\label{eq:query}
\M{q} = f_{q}(\M{x}; \theta_{q}),
\end{equation}
where $\M{q} \in \mathbb{R}^{h}$ denotes the query vector, $h < d$ refers to the dimension of query embeddings, and $f_{q}(\cdot; \theta_{q})$ is given by fully-connected networks. By using Eq.~\eqref{eq:query}, the query encoder is formulated by concatenating $f_{q}$ on top of $f_{cnn}$. In this work, the memory and query encoder share the same CNN weights, and freeze $\theta_{cnn}$ with the pre-trained TSN model to avoid overfitting. Thus, the encoding component of AMemNet is mainly parameterized by $\theta_{enc}=\{\theta_{q}\}$.

\subsubsection{Memory Generator}
We adopt the \emph{key-value} memory network architecture~\cite{KvMemNet,KVMemNet-WWW17} and develop it as a generator model by
\begin{equation}\label{eq:G-Mem}
\M{\hat{v}} = G_{mem}(\M{q}; \theta_{mem}),
\end{equation}
where $G_{mem}(\cdot; \theta_{mem})$ denotes the memory generator and $\M{\hat{v}} \in \mathbb{R}^{d}$ represents the generated full video representation. Particularly, $G_{mem}(\cdot; \theta_{mem})$ includes two memory blocks, termed as a key memory matrix $\M{M}^{k} \in \mathbb{R}^{N \times h}$ and a value memory matrix $\M{M}^{v} \in \mathbb{R}^{N \times d}$, where $N$ is the number of memory slots in each memory block. The memory slot, in essence, is one row in $\M{M}^{k}$/$\M{M}^{v}$, learned with query $\M{q}$ and full video memory $\M{v}$ during the training process. The benefits of using such a key-value structure lies in separating the learning process for different purposes -- $\M{M}^{k}$ could focus on memorizing different queries of partial videos and $\M{M}^{v}$ is trained to distill useful information from full videos for generation. To generate $\M{\hat{v}}$, our memory generator $G_{mem}$ conducts the following three steps in a sequence.

\emph{1) Memory Addressing}. The key memory matrix $\M{M}^{k}$ of $G_{mem}$ provides sufficient flexibility to store similar queries (partial videos) for addressing the relevant value memory slots in $\M{M}^{k}$ with querying attentions. The addressing process is computed by
\begin{small}
\begin{equation}\label{eq:mem-address}
\alpha[i] = \textup{softmax}(\phi(\M{q}, \M{M}^{k}[i])) = \frac{\exp(\phi(\M{q}, \M{M}^{k}[i]))}{\sum_{j=1}^{N} \exp(\phi(\M{q}, \M{M}^{k}[i]))},
\end{equation}
\end{small}where $\alpha \in \mathbb{R}^{N}$ denotes the soft attention weights over all the memory slots, $\M{M}^{k}[i]$ refers to its $i$-th row, and $\phi(\cdot, \cdot)$ is a similarity score function, which could be given by the cosine similarity $\phi(\M{a}, \M{b}) = \M{a}^{\T}\M{b}$ or $\ell_2$ norm $\phi(\M{a}, \M{b}) = -\|\M{a}-\M{b}\|$. Notably, Eq.~\eqref{eq:mem-address} enables an end-to-end differentiable property~\cite{end2endMemNet,KvMemNet} of our memory networks, optimizing the key slots with backpropagation gradients.

\emph{2) Memory Writing}. The value memory matrix $\M{M}^{v}$ of $G_{mem}$ memorizes full videos for the generation purpose, where the memory slots attended by a partial video query $\M{q}$ are written with its full video representation $\M{v}$. Specifically, $G_{mem}$ updates the value memory matrix with gate mechanism and attentions following~\cite{NTM-GravesWD14,KvMemNet}. Let $t$ be the current training step and $\M{M}_{t-1}^{v}$ be the value memory matrix in the last step, 
$\M{M}_{t}^{v} \leftarrow\M{M}_{t-1}^{v}$ is obtained by
\begin{align}
\M{e}_{t} & = \textup{sigmoid}(\M{W}_{e} \M{v})\label{eq:earase},\\ 
\M{a}_{t} & = \textup{tanh}(\M{W}_{a} \M{v})\label{eq:add},\\
\M{\tilde{M}}^{v}_{t}[i] & = \M{M}^{v}_{t-1}[i] \odot (\M{1} - \alpha_{t}[i]\M{e}_{t})\label{eq:Mt-1},\\ 
\M{M}^{v}_{t}[i] & = \M{\tilde{M}}^{v}_{t}[i] + \alpha_{t}[i]\M{a}_t,\label{eq:Mt}
\end{align}
where $\M{e}_{t} \in \mathbb{R}^{d}$ and $\M{a}_{t} \in \mathbb{R}^{d}$ represent the \emph{erase vector} and \emph{add vector}, respectively, $\odot$ denotes the element-wise multiplication, and $\alpha_{t}$ is computed by Eq~\eqref{eq:mem-address} with ($\M{q}$, $\M{v}$) arriving at the $t$-th training step. In Eqs.~\eqref{eq:earase} and~\eqref{eq:add}, the erase vector $\M{e}_{t}$ and add vector $\M{a}_{t}$ work as input and forget gates in the LSTM model~\cite{hochreiter1997long}, implemented by two linear projection matrices\footnote{We omit all the bias vectors to simplify notations.} $\M{W}_{e} \in \mathbb{R}^{d \times d}$ and $\M{W}_{a} \in \mathbb{R}^{d \times d}$, respectively. The $\M{e}_{t}$ decides the forgetting degree of memory slots in $\M{M}^{v}_{t-1}$, while $\M{a}_{t}$ computes the update in $\M{M}^{v}_{t}$. By using query attentions $\alpha_{t}$, Eqs.~\eqref{eq:Mt-1} and~\eqref{eq:Mt} will mainly update the most attended ($\alpha_t[i] \rightarrow 1$) memory slots and leave the ones ($\alpha_t[i] \rightarrow 0$) that are irrelevant to the query $\M{q}$ nearly unchanged.
% TODO: long-term temporal information

\emph{3) Memory Reading}. After updating the value memory matrix $\M{M}^{v}$, $G_{mem}$ generates the full video representation $\M{\hat{v}}$ by reading memory slots from $\M{M}^{v}$ in the following way:
\begin{equation}\label{eq:mem-read}
\M{\hat{v}} = \M{x} + \sum_{i} \alpha[i] \M{M}^{v}[i], 
\end{equation}
which adds a skip-connection between the partial video feature $\M{x}$ and the memory output. Notably, Eq.~\eqref{eq:mem-read} enables $G_{mem}$ to memorize the residual between a partial video and its corresponding entire one.

In summary, the memory generator $G_{mem}(\cdot; \theta_{mem})$ defined in Eq.~\eqref{eq:G-Mem} is implemented through Eq.~\eqref{eq:mem-address} to Eq.~\eqref{eq:mem-read}, where $\theta_{mem}$ includes the key/value memory matrix and all the learnable gate parameters, \emph{i.e.}, $\theta_{mem} = \{\M{M}^{k}, \M{M}^{v}, \M{W}_{e}, \M{W}_{a}\}$.

\subsubsection{Discriminator} The discriminator network is designed with two purposes as 1) predicting the true action category label given the real/generated ($\M{v}$/$\M{\hat{v}}$) full video representation, and 2) distinguishing the real full video representation $\M{v}$ and the fake one $\M{\hat{v}}$. Inspired by~\cite{StarGAN,AAPnet-TPAMI20}, we build the discriminator in a composition way: $D(\cdot; \theta_{D}):=\{D_{cls}(\cdot; \theta_{cls}), D_{adv}(\cdot; \theta_{adv})\}$, where $D_{cls}: \mathbb{R}^{d} \rightarrow \mathbb{R}^{|\C{Y}|}$ works as a classifier to predict probability scores over $|\C{Y}|$ action classes, and $D_{adv}: \mathbb{R}^{d} \rightarrow \{0,1\}$ follows the same definition in the GAN model~\cite{GoodfellowNIPS2014} to infer the probability of the given sample being real. The discriminator $D$ in our model is formulated as fully-connected networks parameterized by $\theta_{D} = \{\theta_{cls}, \theta_{adv}\}$.

\subsection{Objective Function} The main goal of this work is to deliver realistic full-video representations for partial videos to predict their correct action classes. To this end, three loss functions are jointly employed for training the proposed AMemNet model.

\textbf{Adversarial Loss}. Given a partial video feature $\M{x}$ and its real full video representation $\M{v}$, we compute the adversarial loss $\C{L}_{adv}$ by
\begin{small}
\begin{equation}\label{eq:loss-adv}
\C{L}_{adv} = \mathbb{E}_{\M{v}} [\log D_{adv}(\M{v})] + \mathbb{E}_{\M{q}} [\log (1 - D_{adv}(G_{mem}(\M{q})))].
\end{equation}
\end{small}The discriminator $D_{adv}$ tries to differentiate $\M{\hat{v}} = G_{mem}(\M{q})$ from the real one $\M{v}$ by maximizing $\C{L}_{adv}$, while, on the contrary, the memory generator $G_{mem}$ aims to fool $D_{adv}$ by minimizing $\C{L}_{adv}$. By using Eq.~\eqref{eq:loss-adv}, we could employ $D_{adv}$ to push $G_{mem}$ towards generating realistic full video features.

\textbf{Reconstruction Loss}. The adversarial loss $\C{L}_{adv}$ encourages our model to generate video features approaching the real feature distribution of full videos, yet without considering the reconstruction error at an instance level, which might miss some useful information for recovering $\M{v}$ from $\M{x}$. In light of this, we define a reconstruction loss as
\begin{equation}\label{eq:loss-rec}
\C{L}_{rec} =\mathbb{E}_{(\M{x}, \M{v})} \|G_{mem}(f_q(\M{x}))  - \M{v}\|_{2}^{2},
\end{equation}
which calculates the squared Euclidean distance between the generated feature $\M{\hat{v}} = G_{mem}(f_q(\M{x}))$ and its corresponding full video feature $\M{v}$. Eq.~\eqref{eq:loss-rec} further guides the memory generator by bridging the gap between $\M{x}$ and $\M{v}$.

\textbf{Classification Loss}. It is important for $G_{mem}$ to generate discriminative representations $\M{\hat{v}}$ for predicting different action classes. Thus, it is natural to impose a classification loss $\C{L}_{cls}$ on training the memory generator as follows:
\begin{align}
\C{L}_{cls}^{v} &= \mathbb{E}_{(\M{v}, \M{y})} H(\M{y}, D_{cls}(\M{v})),\label{eq:loss-cls-D}\\
\C{L}_{cls}^{x} &= \mathbb{E}_{(\M{x}, \M{y})} H(\M{y}, D_{cls}(G_{mem}(\M{x}))),\label{eq:loss-cls-G}
\end{align}
where $\M{y} \in \mathbb{R}^{|\C{Y}|}$ indicates the one-hot vector of an action label $y$ over $|\C{Y}|$ classes and $H(\cdot, \cdot)$ computes the cross-entropy between two probability distributions. Let $\M{\hat{y}} \in \mathbb{R}^{|\C{Y}|}$ be the output of $D_{cls}$, we have 
$H(\M{y}, \M{\hat{y}}) = -\sum_{i=1}^{|\C{Y}|}\M{y}[i]\log\M{\hat{y}}[i]$.

Different from~\cite{AAPnet-TPAMI20}, which only employs the classification loss $\C{L}_{cls}^{v}$ in training the discriminator model with full videos, we employ Eq.~\eqref{eq:loss-cls-D} and Eq.~\eqref{eq:loss-cls-G} to train the discriminator $D_{cls}$ and the memory generator $G_{mem}$ alternatively. The benefit lies in: a high-quality classifier $D_{cls}$ is first obtained by minimizing $\C{L}_{cls}^{v}$ with real full videos and then $D_{cls}$ is leveraged to ``teach'' $G_{mem}$ for generating representations $\M{\hat{v}}$ to lower $\C{L}_{cls}^{x}$. By this means, $G_{mem}$ could learn the discriminative information from $D_{cls}$.

\textbf{Final Objective}. By summarizing Eq.~\eqref{eq:loss-adv} to Eq.~\eqref{eq:loss-cls-G}, the final objective function of the proposed AMemNet model is given by
\begin{align}
\max_{\theta_{D}} &~ \C{L}_{adv} + \lambda_{cls} \C{L}_{cls}^{v}, \label{eq:loss-D}\\
\min_{\theta_{G}} &~ \C{L}_{adv} + \lambda_{cls} \C{L}_{cls}^{x} + \lambda_{rec} \C{L}_{rec}, \label{eq:loss-G}
\end{align}
where $\theta_{G}=\{\theta_{enc}, \theta_{mem}\}$ includes all the trainable parameters for generating $\M{\hat{v}}$ from $\M{x}$, $\theta_{D} = \{\theta_{cls}, \theta_{adv}\}$ parametrizes the discriminator, and $\lambda_{cls}$ and $\lambda_{rec}$ are the trade-off parameters balancing different loss functions. To proceed with the training procedure, we optimize $\theta_{D}$ and $\theta_{G}$ by alternatively solving Eq.~\eqref{eq:loss-D} and Eq.~\eqref{eq:loss-G} while fixing the other.

\begin{figure}[t]
\centering
\includegraphics[width=\linewidth]{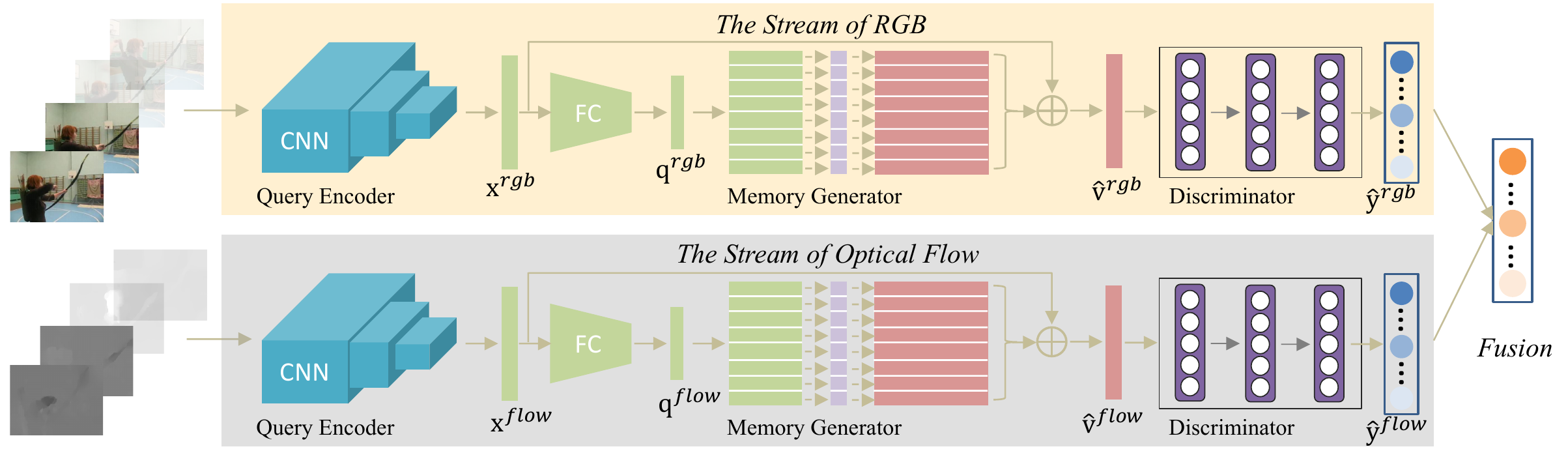}
\caption{Illustration of applying AMemNets over RGB and optical flows.}\label{fig:fusion}\vspace{-0.2cm}
\end{figure}

\subsection{Two-Stream Fusion for Action Prediction}
After training the proposed AMemNet model via Eq.~\eqref{eq:loss-D} and Eq.~\eqref{eq:loss-G}, we freeze the model weights $\theta = \{\theta_{G}, \theta_{D}\}$ and suppress the memory writing operations in Eq.~\eqref{eq:earase}-\eqref{eq:Mt} for testing AMemNet. Particularly, given a partial video feature $\M{x}$, we predict its action label by
\begin{equation}\label{eq:yhat}
\begin{aligned}
\M{\hat{y}} = D_{cls}(G_{mem}(f_{q}(\M{x}))),
\end{aligned}
\end{equation}
where $\M{\hat{y}} \in \mathbb{R}^{|\C{Y}|}$ denotes the probability distribution over $|\C{Y}|$ action classes for $\M{x}$.

As shown in Fig~\ref{fig:fusion}, we adopt a two-stream framework~\cite{SimonyanNIPS2014,WangECCV2016} to exploit the spatial and temporal information of given videos, where we first test AMemNet on each stream (i.e., RGB frames and optical flow) individually, and then fuse the prediction scores to obtain the final result. Given $\M{x}^{rgb}$ and $\M{x}^{flow}$, we obtain the prediction results $\M{\hat{y}}^{rgb}$ and $\M{\hat{y}}^{flow}$ by using Eq.~\eqref{eq:yhat} with $\theta^{rgb}$ and $\theta^{rgb}$, respectively. The final prediction result is given by
\begin{equation}\label{eq:yfusion}
\M{\hat{y}}^{fusion} = \M{\hat{y}}^{rgb} + \beta \M{\hat{y}}^{flow},
\end{equation}
where $\beta$ is the fusion weight for integrating the scores given by the stream of spatial RGB frames and the stream of temporal optical flow images.

\begin{table*}[t]
\begin{center}
\caption{Action prediction accuracy (\%) under 10 observation ratios on the UCF101 dataset.}\label{tab:ucf101}%\vspace{-0.2cm}
\scalebox{0.9}{
\begin{tabular}{llcccccccccc}
  \toprule
& Method & $0.1$ & $0.2$ & $0.3$ & $0.4$& $0.5$ & $0.6$ & $0.7$ & $0.8$ & $0.9$& $1.0$\\
\midrule
\multirow{6}{*}{\emph{Single-stream}}
& IBoW \cite{RyooICCV2011}        & $36.29$ & $65.69$ & $71.69$ & $74.25$ & $74.39$ & $75.23$ & $75.36$ & $75.57$ & $75.79$ & $75.79$\\
& MSSC \cite{CaoCVPR2013}         & $34.05$ & $53.31$ & $58.55$ & $57.94$ & $61.79$ & $60.86$ & $63.17$ & $63.64$ & $61.63$ & $61.63$\\
& MTSSVM \cite{KongECCV2014}      & $40.05$ & $72.83$ & $80.02$ & $82.18$ & $82.39$ & $83.21$ & $83.37$ & $83.51$ & $83.69$ & $82.82$\\
& DeepSCN~\cite{KongCVPR2017}     & $45.02$ & $77.64$ & $82.95$ & $85.36$ & $85.75$ & $86.70$ & $87.10$ & $87.42$ & $87.50$ & $87.63$\\
& PA-DRL~\cite{Chen_2018_ECCV}            & $81.36$ & $82.63$ & $82.90$ & $83.51$ & $84.01$ & $84.38$ & $85.09$ & $85.41$ & $85.81$ & $86.15$\\
& PTSL~\cite{Wang_2019_CVPR}              & $83.32$ & $87.13$ & $88.92$ & $89.82$ & $90.85$ & $91.04$ & $91.28$ & $91.23$ & $91.31$ & $91.47$\\
\midrule
\multirow{4}{*}{\emph{Two-stream}}
& Mem-LSTM~\cite{KongAAAI2018}          & $51.02$ & $80.97$ & $85.73$ & $87.76$ & $88.37$ & $88.58$ & $89.09$ & $89.38$ & $89.67$ & $90.49$ \\
& TSL~\cite{TSL-wacv18}                     & $82.20$ & $86.70$ & $88.50$ & $89.50$ & $90.10$ & $91.00$ & $91.50$ & $91.90$ & $92.40$ & $92.50$ \\
& RGN-KF~\cite{Zhao_2019_ICCV}              & $83.12$ & $85.16$ & $88.44$ & $90.78$ & $91.42$ & $92.03$ & $92.00$ & $93.19$ & $93.13$ & $93.14$ \\
& AAPNet~\cite{AAPnet-TPAMI20}        & $90.25$ & $93.10$ & $94.46$ & $95.41$ & $95.89$ & $96.09$ & $96.27$ & $96.35$ & $96.47$ & $96.36$ \\
\midrule
\multirow{3}{*}{\emph{Baselines}}
& TSN~\cite{WangECCV2016}           & $86.76$ & $89.29$ & $90.64$ & $91.81$ & $91.73$ & $92.47$ & $92.97$ & $93.15$ & $93.31$ & $93.42$ \\
& TSN+finetune                & $88.88$ & $91.52$ & $93.01$ & $94.05$ & $94.66$ & $95.34$ & $95.64$ & $95.92$ & $95.90$ & $96.00$ \\
& TSN+KNN                   & $85.69$ & $88.71$ & $90.34$ & $91.29$ & $91.78$ & $92.33$ & $92.41$ & $92.75$ & $92.96$ & $93.11$ \\
\midrule
\multirow{3}{*}{\emph{Our model}}
& AMemNet-RGB                 & $85.95$ & $87.47$ & $88.21$ & $88.57$ & $89.26$ & $89.51$ & $89.81$ & $89.99$ & $90.06$ & $90.17$ \\
& AMemNet-Flow                & $83.64$ & $88.32$ & $90.74$ & $92.18$ & $93.20$ & $93.78$ & $94.40$ & $94.75$ & $94.88$ & $94.96$ \\
& AMemNet                   & $\pmb{92.45}$ & $\pmb{94.60}$ & $\pmb{95.55}$ & $\pmb{96.00}$ & $\pmb{96.45}$ & $\pmb{96.67}$ & $\pmb{96.97}$ & $\pmb{96.95}$ & $\pmb{97.07}$ & $\pmb{97.03}$ \\
\bottomrule
\end{tabular}}\vspace{-0.4cm}
\end{center}
\end{table*}

\section{Experiments}
\subsection{Experimental Setting}
\textbf{Datasets}. Two benchmark video datasets, UCF101~\cite{Soomro2012} and HMDB51~\cite{HMDB51}, are used in the experiment. The UCF101 dataset consists of $13,320$ videos from $101$ human actions covering a wide range of human activities, and the HMDB51 dataset collects $6,766$ video clips from movies and web videos over $51$ action categories. We follow the standard training/testing splits on these two datasets following~\cite{WangECCV2016,AAPnet-TPAMI20}. We test the proposed AMemNet model over \emph{three splits} and report the average prediction result for each dataset. We employ the preprocessed RGB frames and optical flow images provided in~\cite{Feichtenhofer_2016_CVPR}.

\textbf{Implementation Details}. The proposed AMemNet is built on top of temporal segment networks (TSN)~\cite{WangECCV2016}, where we adopt the BN-Inception network~\cite{BN-icml15} as its backbone and employ the pre-trained model on the Kinetics dataset~\cite{Kinetics}. The same data augmentation strategy (\emph{e.g.}, cropping and jittering) as provided in~\cite{WangECCV2016} is employed for encoding all the partial and full videos as $d=1024$ feature representations. We formulate $f_{q}$ as fully-connected networks of two layers, where the middle layer has $512$ hidden states and the final query embedding size is set as $h=256$. The batch normalization and LealyRelu are both used in $f_{q}$. We employ $N=512$ memory slots for the key and value memory matrices, hence we have $\M{M}^{k} \in \mathbb{R}^{512 \times 256}$ and $\M{M}^{v} \in \mathbb{R}^{512 \times 1024}$. All the memory matrices and gating parameters in $\theta_{mem}$ are randomly initialized. We implement the discriminator network by one fully-connected layer, where the softmax and sigmoid activation function are used for $D_{cls}$ and $D_{adv}$, respectively. 

For each training step, we first employ the Adam optimizer with a learning rate of $0.0001$ to update $\theta_{D}$ with Eq.~\eqref{eq:loss-D} twice, and then optimize $\theta_{G}$ once by solving Eq.~\eqref{eq:loss-G} with the SGD optimizer of $0.0001$ learning rate and $0.9$ momentum rate. We set the batch size as $64$. For all the datasets, we set $\lambda_{cls}=1$ to strengthen the impact of $D_{cls}$ on the memory generator $G_{mem}$ for encouraging discriminative representations, and set $\lambda_{res}=0.1$ to avoid overemphasizing the reconstruction of each video sample to lead the overfitting issue. The fusion weight $\beta$ is fixed as $1.5$ for all the datasets following~\cite{WangECCV2016}. All the codes in this work were implemented by Pytorch and ran with Titan X GPUs.

\textbf{Compared Methods}. We compare our approach with three different kinds of methods as follows. 
1) \emph{Single-stream} methods: Integral BoW (IBoW)~\cite{RyooICCV2011}, mixture segments sparse coding (MSSC)~\cite{CaoCVPR2013}, multiple temporal scales SVM (MTSSVM)~\cite{KongECCV2014}, deep sequential context networks (DeepSCN)~\cite{KongCVPR2017}, part-activated deep reinforcement learning (PA-DRL)~\cite{Chen_2018_ECCV}, and progressive teacher-student learning (PTSL)~\cite{Wang_2019_CVPR}.
2) \emph{Two-stream} methods: memory augmented LSTM (Mem-LSTM)~\cite{KongAAAI2018}, temporal sequence learning (TSL)~\cite{TSL-wacv18}, residual generator network with Kalman filter (RGN-KF)~\cite{Zhao_2019_ICCV}, and adversarial action prediction networks (AAPNet)~\cite{AAPnet-TPAMI20}. We implemented the AAPNet with the same pre-trained TSN features as our approach and posted the authors' reported results for the other single/two-stream methods.
3) \emph{Baselines}: We also compare AMemNet with temporal segment networks (TSN)~\cite{WangECCV2016}. Specifically, we test the TSN model pre-trained on the UCF101/HMDB51 dataset as baseline results, and finetune the TSN model pre-trained on the Kinetics dataset for UCF101 and HMDB51, respectively. Moreover, we train a k-nearest neighbors (KNN) classifier with the Kinetics pre-trained TSN features, termed as TSN+KNN, and report its best performance by selecting $k$ from $\{5,10,20,30,50,100,500\}$. For a fair comparison, we follow the same testing setting in previous works~\cite{KongCVPR2017,Chen_2018_ECCV,Zhao_2019_ICCV} by evenly dividing all the videos into 10 progresses, \emph{i.e.}, $P=10$ as described in Section~\ref{sec:setting}. However, it is worth noting that, the proposed AMemNet does not require any progress label in both training and testing.

\begin{table*}[t]
\begin{center}
\caption{Action prediction accuracy (\%) under 10 observation ratios on the HMDB51 dataset.}\label{tab:hmdb51}
\scalebox{0.9}{
\begin{tabular}{llcccccccccc}
  \toprule
& Method & $0.1$ & $0.2$ & $0.3$ & $0.4$& $0.5$ & $0.6$ & $0.7$ & $0.8$ & $0.9$& $1.0$\\
\midrule
\multirow{3}{*}{\emph{Two-stream}}
& Global-Local~\cite{GL-TDM-TIP18}              & $38.80$   & $43.80$   & $49.10$   & $50.40$   & $52.60$   & $54.70$   & $56.30$   & $56.90$   & $57.30$   & $57.30$ \\
& TSL~\cite{TSL-wacv18}                         & $38.80$   & $51.60$   & $57.60$   & $60.50$   & $62.90$   & $64.60$   & $65.60$   & $66.20$   & $66.30$   & $66.30$ \\
& AAPNet~\cite{AAPnet-TPAMI20}          & $56.03$   & $60.11$ & $64.87$   & $67.99$   & $70.76$   & $72.55$   & $74.00$   & $74.81$   & $75.59$   & $75.56$ \\
\midrule
\multirow{3}{*}{\emph{Baselines}}
& TSN~\cite{WangECCV2016}             & $47.12$   & $52.81$   & $59.35$   & $62.55$   & $64.77$   & $67.52$   & $68.95$   & $69.87$   & $70.07$   & $70.13$ \\
& TSN+finetune                  & $55.13$   & $59.82$ & $63.88$   & $67.02$   & $69.74$   & $71.72$   & $72.98$   & $73.43$   & $74.08$   & $73.55$ \\
& TSN+KNN                     & $48.77$   & $53.59$ & $57.83$   & $60.33$   & $62.84$   & $65.18$   & $66.53$   & $67.00$   & $67.53$   & $66.65$ \\
\midrule
\multirow{3}{*}{\emph{Our model}}
& AMemNet-RGB     & $52.55$ & $55.52$ & $58.27$ & $60.55$ & $62.53$ & $63.87$ & $64.41$ & $64.61$ & $64.99$ & $64.86$ \\
& AMemNet-Flow    & $47.41$ & $54.43$ & $60.26$ & $64.51$ & $68.03$ & $70.53$ & $72.10$ & $73.05$ & $73.39$ & $73.52$ \\
& AMemNet         & $\pmb{57.74}$ & $\pmb{62.10}$ & $\pmb{66.28}$ & $\pmb{70.17}$ & $\pmb{72.66}$ & $\pmb{74.55}$ & $\pmb{75.22}$ & $\pmb{75.78}$ & $\pmb{76.08}$ & $\pmb{76.14}$ \\
\bottomrule
\end{tabular}}\vspace{-0.4cm}
\end{center}
\end{table*}

\begin{figure*}[ht]
     \subfigure[RGB on UCF101]{\label{fig:ucf101-rgb}
      \begin{minipage}[c]{0.31\textwidth}
      \centering
      \includegraphics[width=1\textwidth]{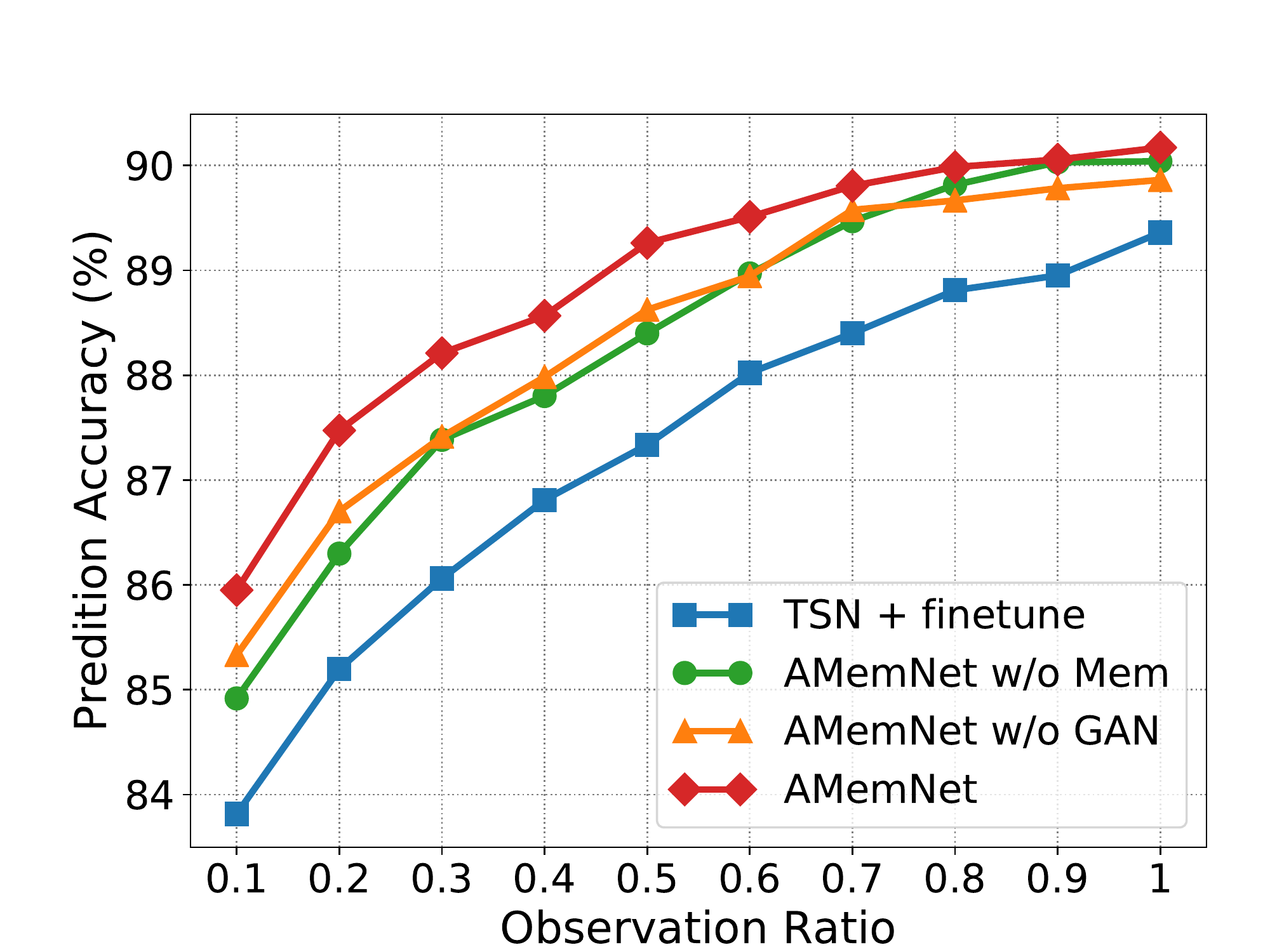}\vspace{0.1cm}
      \end{minipage}}
     \hfill
     \subfigure[Flow on UCF101]{\label{fig:ucf101-flow}
      \begin{minipage}[c]{0.32\textwidth}
      \centering
      \includegraphics[width=1\textwidth]{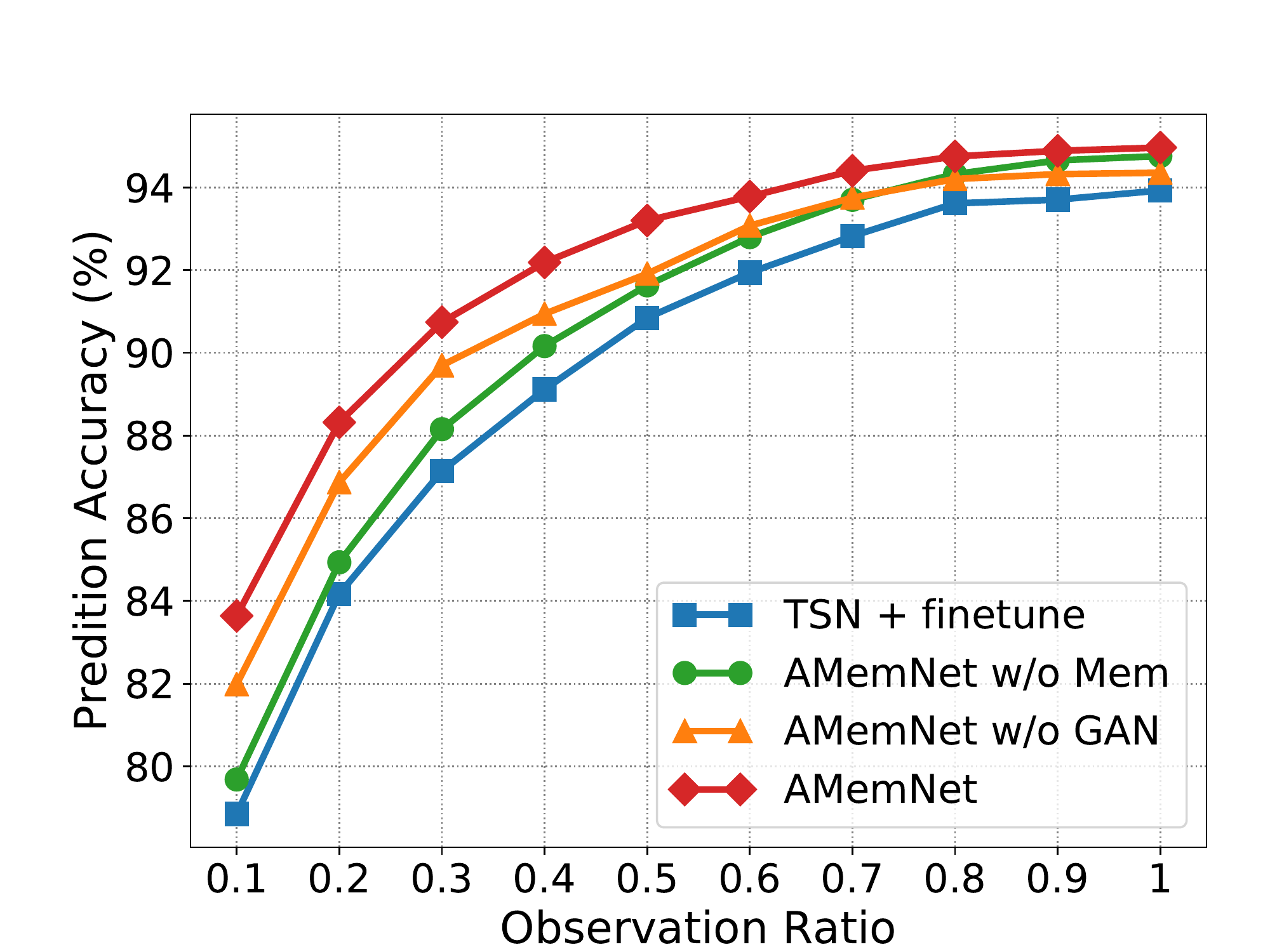}\vspace{0.1cm}
      \end{minipage}}
     \hfill
     \subfigure[Fusion on UCF101]{\label{fig:ucf101-fusion}
      \begin{minipage}[c]{0.32\textwidth}
      \centering
      \includegraphics[width=1\textwidth]{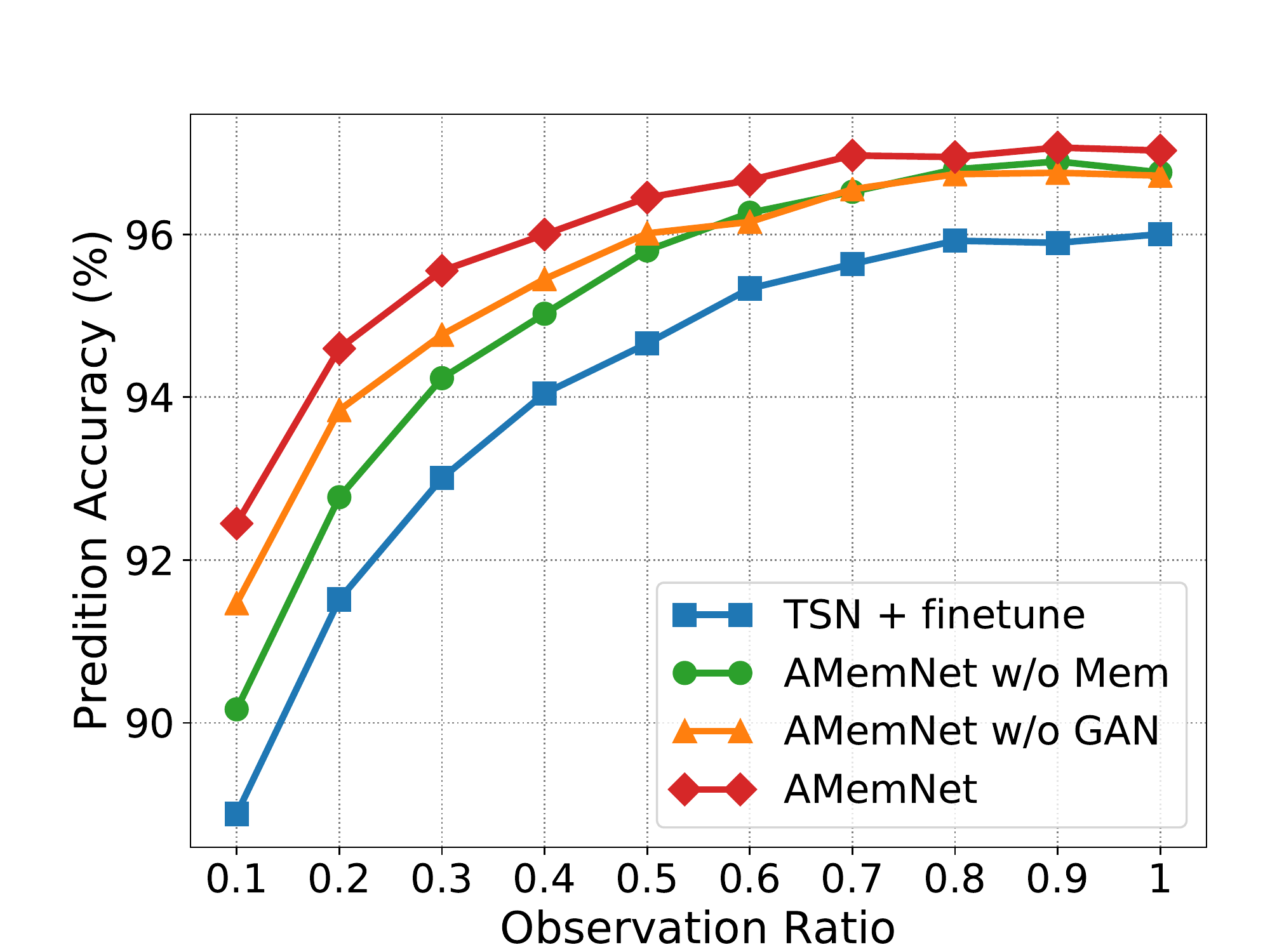}\vspace{0.1cm}
      \end{minipage}}\vspace{-0.3cm}
    \caption{Ablation study for the proposed AMemNet on the UCF101 dataset in terms of RGB, Flow and Fusion, respectively.}\label{fig:ablation}
\end{figure*}

\begin{figure*}[ht]
     \subfigure[Predictions accuracy]{\label{fig:atts}
        \begin{minipage}[c]{0.32\textwidth}
          \centering
          \includegraphics[width=\textwidth]{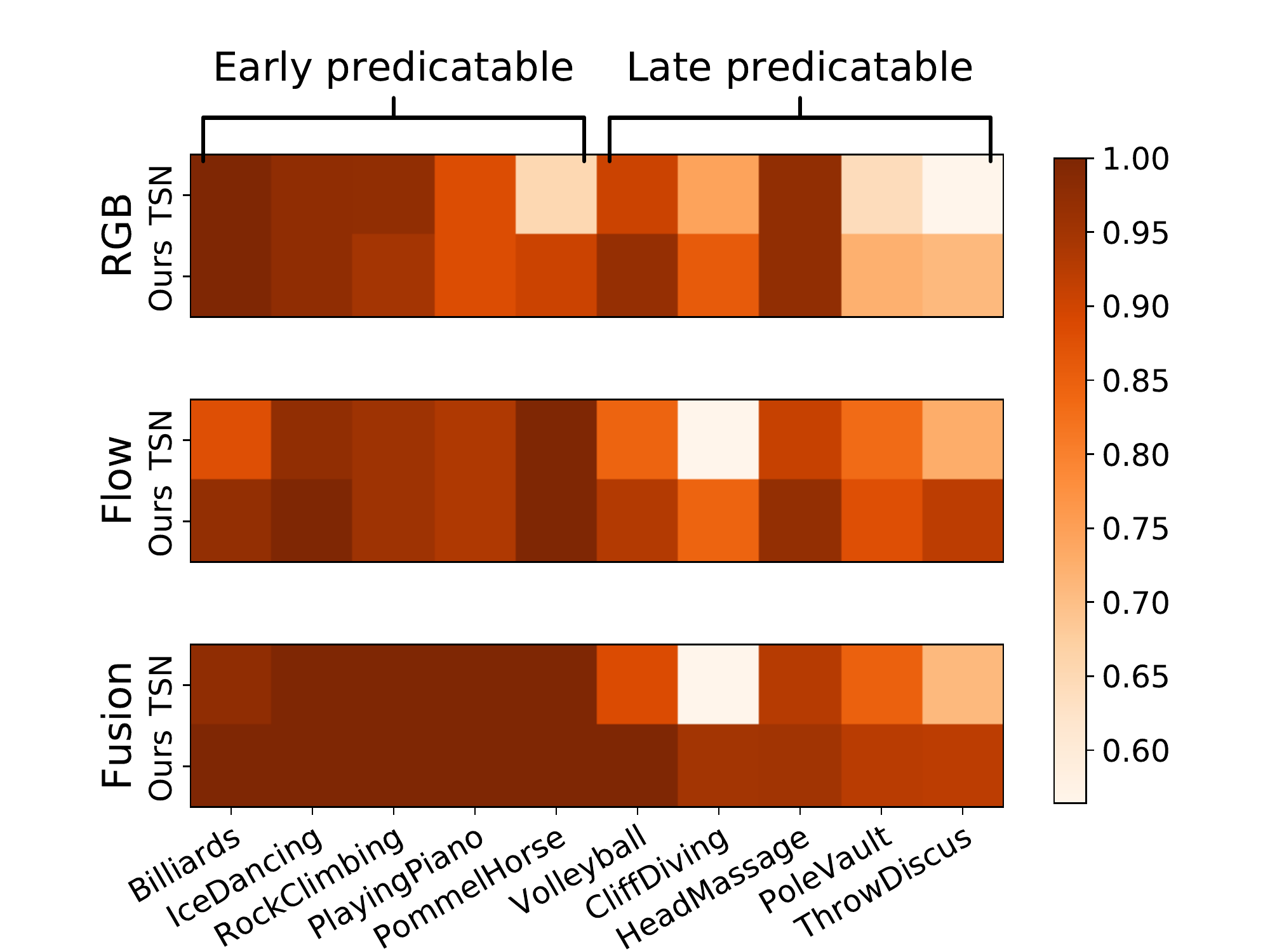}
        \end{minipage}}
     \hfill
     \subfigure[TSN]{\label{fig:tnse-base}
        \begin{minipage}[c]{0.32\textwidth}
          \centering
          \includegraphics[width=\textwidth]{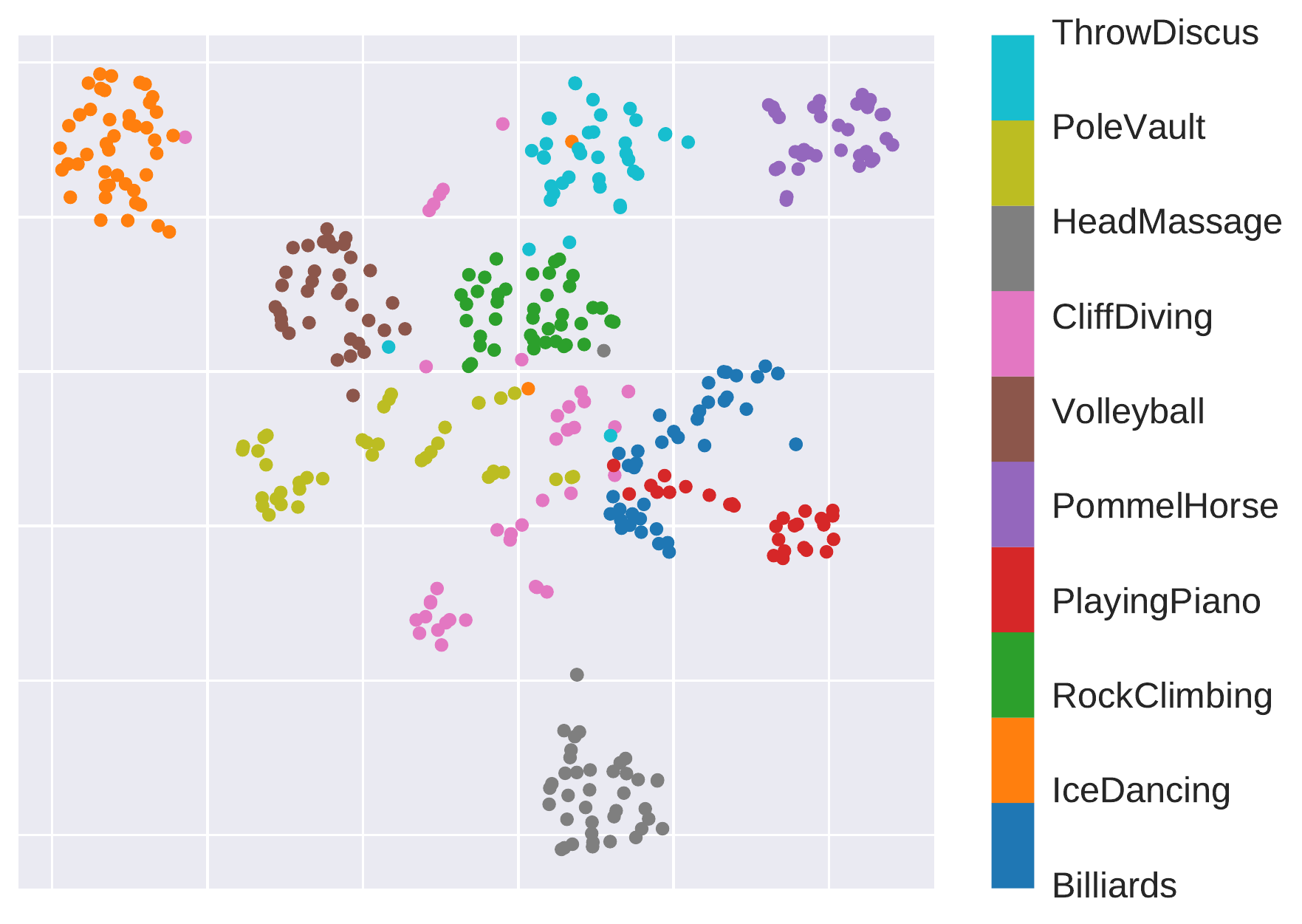}
        \end{minipage}}
     \hfill
     \subfigure[AMemNet]{\label{fig:tsn-KVMem}
        \begin{minipage}[c]{0.32\textwidth}
          \centering
          \includegraphics[width=\textwidth]{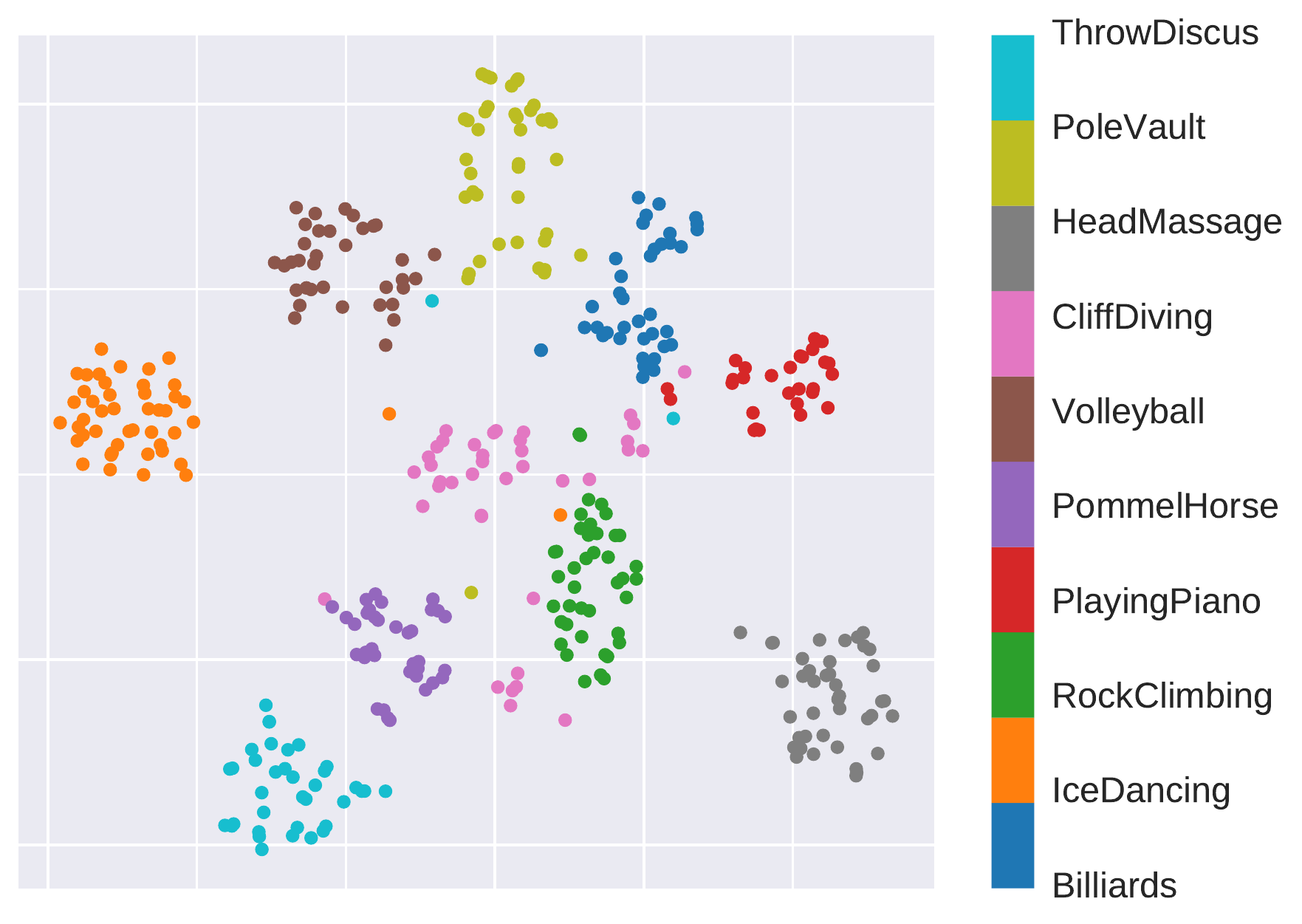}
        \end{minipage}}\vspace{-0.2cm}
     \caption{Visual analysis of the proposed AMemNet between \emph{early predictable} and \emph{late predictable} action categories. (a) Prediction results of TSN and the proposed AMemNet over 10 categories with $10\%$ video progress on UCF101. (b) and (c) t-SNE embedding results given by TSN and AMemNet.}\label{fig:vis}\vspace{-0.2cm}
\end{figure*}

\subsection{Prediction Performance}
\textbf{UCF101 Dataset}.
Table~\ref{tab:ucf101} summarizes the prediction accuracy of the proposed AMemNet and 13 compared methods on the UCF101 dataset. Overall, AMemNet consistently outperforms all the competitors over different observation ratios with a significant improvement. Impressively, the proposed AMemNet achieves around $92\%$ accuracy when only $10\%$ video is observed, which fully validates the effectiveness of applying AMemNet for early action prediction. This is mainly benefited from the rich \emph{key-value} structured memories learned from full-video features guided by the adversarial training.

The single-stream methods mainly explore the temporal information by using hand-crafted features (\emph{e.g.}, spatio-temporal interest points (STIP)~\cite{Dollar2005}, dense trajectory) like in IBoW~\cite{RyooICCV2011}, MSSC~\cite{CaoCVPR2013}, and MTSSVM~\cite{KongECCV2014}, or by utilizing 3D convolutional networks (\emph{e.g.}, C3D~\cite{TranICCV2015}) like in DeepSCN~\cite{KongCVPR2017} and PTSL~\cite{Wang_2019_CVPR}. Differently, the two-stream methods deploy the convolutional neural networks on two pathways to capture the spatial information of RGB images and the temporal characteristics of optic flows, respectively. On the one hand, the two-stream methods could better exploit the spatial-temporal information inside videos than using one single stream. The proposed AMemNet inherits the merits from this two-stream architecture, and thus performs better than the single-stream methods that even employ a more powerful CNN encoder, \emph{e.g.}, the 3D ResNeXt-101~\cite{Hara_2018_CVPR} used in PTSL~\cite{Wang_2019_CVPR} is much deeper than BN-Inception in the proposed AMemNet. On the other hand, compared with two-stream methods, especially the AAPNet~\cite{AAPnet-TPAMI20} implemented with the same backbone as our model, the consistent improvement of AMemNet over AAPNet shows the effectiveness of using the memory generator to deliver ``full'' video features in early progress.

In Table~\ref{tab:ucf101}, we refer to AMemNet-RGB and AMemNet-Flow as the single-stream result by using AMemNet on RGB frames and flow images, respectively. Two interesting observations could be drawn: 1) The RGB contributes more than the flow at the beginning, as the still images encapsulating scenes and objects could provide key clues for recognizing the actions with few frames. 2) The late fusion naturally fits action prediction by integrating the complimentary information between two streams over time.

\textbf{HMDB51 Dataset}.
Table~\ref{tab:hmdb51} reports the prediction results of our approach and TSN~\cite{WangECCV2016} on the HMDB51 dataset, which, compared with UCF101, is a more challenging dataset for predicting actions due to the large motion variations rather than static cues across different categories~\cite{HMDB51}. As can be seen, the flow result of AMemNet exceeds AMemNet-RGB around $8\%$ accuracy after more progress being observed (\emph{e.g.}, $\tau_{p} \geq 0.5$). However, even under this case, the proposed AMemNet still consistently improves AMemNet-Flow by incorporating RGB results along with different progresses. Moreover, the clear improvements of AMemNet over TSN indicate that the full video memories learned by our memory generator could well enhance the discriminability of video representations in early progress.

\subsection{Model Discussion}
\textbf{Ablation Study}. Fig.~\ref{fig:ablation} shows the ablation study of the proposed AMemNet model on the UCF101 dataset\footnote{
  More ablation study results and parameter analyses on the UCF101 and HMDB51 datasets are provided in the supplementary material.
} in terms of RGB, Flow and fusion, respectively, where we test all the methods by different observation ratios. We adopt TSN + finetune as a sanity check for our approach and implement two strong ablated models to discuss the impact of two main components of AMemNet as follows. 1) \texttt{AMemNet w/o Mem} refers to our model by discarding the memory generator, \emph{i.e.}, $\theta \backslash \theta_{mem}$. Instead, we use the same generator network as in AAPNet~\cite{AAPnet-TPAMI20} for generating full video features with AMemNet w/o Mem. 2) \texttt{AMemNet w/o GAN} is developed without the adversarial training and is trained by only using a classification loss.

As shown in Fig~\ref{fig:ablation}, AMemNet improves all the above methods with a clear margin on different cases, which strongly supports the motivation of this work. 
It is worth noting that, for the early progress (\emph{i.e.}, the observation ratio $\tau_{p} \leq 0.3$), AMemNet w/o GAN clearly boots the performance over AMemNet w/o Mem. This demonstrates the effectiveness of using memory networks to compensate for the limited information in incomplete videos. As observing more progress, the GAN model will lead the generating process since it has sufficient information given by the partial videos, where AMemNet w/o Mem improves over AMemNet w/o GAN after $\tau_{p} > 0.7$ on the UCF101 dataset.

\textbf{Early Predicable vs Late Predicable}. In Fig.~\ref{fig:vis}, we discuss the performance of AMemNet for action categories of different properties, \emph{e.g.}, \emph{predictability} (the progress level required for recognizing an action), on the UCF101 dataset. We compare AMemNet and TSN~\cite{WangECCV2016} on the $10\%$ progress level video of 10 different categories in~\ref{fig:atts} and show the corresponding t-SNE~\cite{t-SNE} embeddings of TSN features and the generated full video features given by AMemNet in~\ref{fig:tnse-base} and~\ref{fig:tsn-KVMem}, respectively. Inspired by~\cite{KongCVPR2017}, we select 10 action categories from UCF101 and divide them into two groups as 1) the early predictable group including \emph{Billiards, IceDancing, RockClimbingIndoor, PlayingPiano, PommelHorse}, and 2) the late predictable group including \emph{VolleyballSpiking, CliffDiving, HeadMassage, PoleVault, ThrowDiscus}, where the early group usually could be predicted by given $10\%$ progress and the late group is selected as the non-early ones.

As expected, the proposed AMemNet mainly improves the TSN baseline over late predictable actions in Fig.~\ref{fig:atts}, which again demonstrates the realistic of the full video features given by our memory generator. Moreover, as shown in Fig.~\ref{fig:tnse-base} and~\ref{fig:tsn-KVMem}, while TSN exhibits a good structured feature embeddings for early predictable classes, \emph{e.g.}, \emph{IceDancing} and \emph{PommelHorse}, its embeddings mixes up for the late predictable ones like \emph{PoleVault} and \emph{CliffDiving}. In contrast, AMemNet generates full video features that encourage a good cluster structure in the embedding space.

\section{Conclusion}
In this paper, we presented a novel two-stream adversarial memory networks (AMemNet) model for the action prediction task. A key-value structured memory generator was proposed to generate the full video feature conditioning on the partial video query, and a class-aware discriminator was developed to supervise the generator for delivering realistic and discriminative representations towards full videos through adversarial training. The proposed AMemNet adopts input and forget gates for updating the full video memories attended by different queries, which captures the long-term temporal variation across different video progresses. Experimental results on two benchmark datasets were provided to demonstrate the effectiveness of AMemNet for the action prediction problem compared with state-of-the-art methods.

\bibliographystyle{abbrv}
%\bibliography{egbib,acmart,egbib_aapnet}
%\bibliography{arxiv}

\appendix

\begin{table}[ht]
\begin{center}
\caption{Overall comparison between ablated models. We provide two-stream fusion results of the $p=0.1$ observation ratio and the \emph{averaged} results over 10 observation ratios. Generally, $p=0.1$ is the most important observation ratio for action prediction.
}\label{tab:abl-mth}\vspace{0.2cm}
\scalebox{1.0}{
\begin{tabular}{lccccccc}
  \toprule
\multirow{2}{*}{Methods} & \multicolumn{3}{c}{Model Components} & \multicolumn{2}{c}{UCF101} & \multicolumn{2}{c}{HMDB51}\\
\cmidrule(lr){2-4}
\cmidrule(lr){5-6}\cmidrule(lr){7-8}
                  & $\C{L}_{rec}$ & $\C{L}_{adv}$ & $\theta_{mem}$ & $0.1$ & \emph{avg.} & $0.1$ & \emph{avg.} \\
\midrule
TSN + finetune          &         &          &         & $88.88$ & $94.09$ & $55.13$ & $68.13$ \\
AMemNet w/o Mem      & \checkmark & \checkmark  &            & $90.17$ & $95.12$ & $55.49$ & $68.68$ \\
AMemNet w/o GAN         & \checkmark &             & \checkmark & $91.47$ & $95.45$ & $56.36$ & $70.30$ \\
AMemNet w/o Res      &         & \checkmark  & \checkmark & $92.26$ & $95.62$ & $56.00$ & $70.33$ \\
AMemNet (ours)          & \checkmark & \checkmark  & \checkmark & $\pmb{92.45}$ & $\pmb{95.97}$ & $\pmb{57.74}$ & $\pmb{70.67}$ \\
\bottomrule
\end{tabular}}
\end{center}
\end{table}

\section{Supplementary Material}\label{app:abl}

\begin{figure*}[ht]
     \subfigure[RGB on HMDB51]{\label{fig:hmdb51-rgb}
      \begin{minipage}[c]{0.32\textwidth}
      \centering
      \includegraphics[width=1\textwidth]{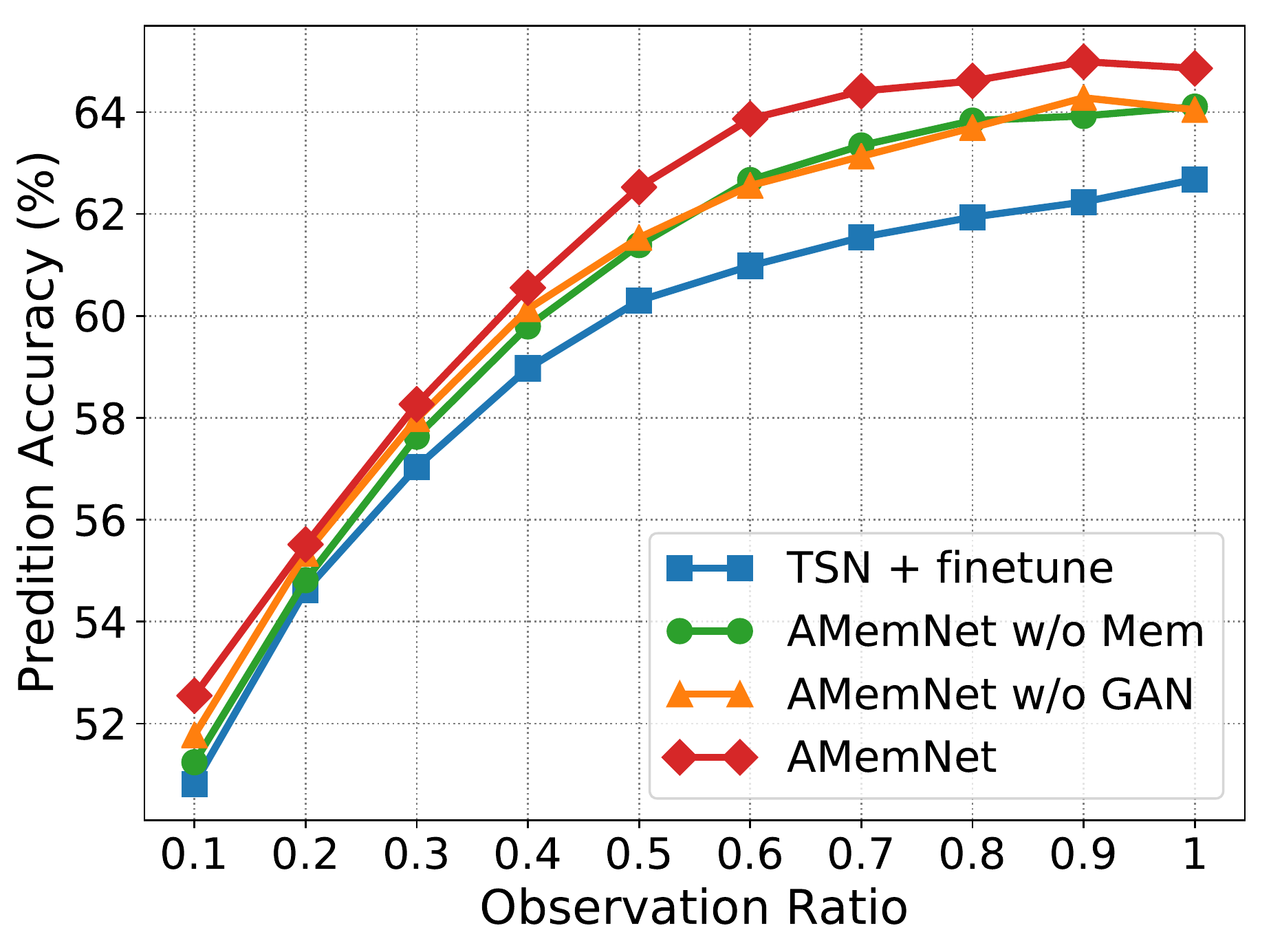}
      \end{minipage}}
     \hfill
     \subfigure[Flow on HMDB51]{\label{fig:hmdb51-flow}
      \begin{minipage}[c]{0.32\textwidth}
      \centering
      \includegraphics[width=1\textwidth]{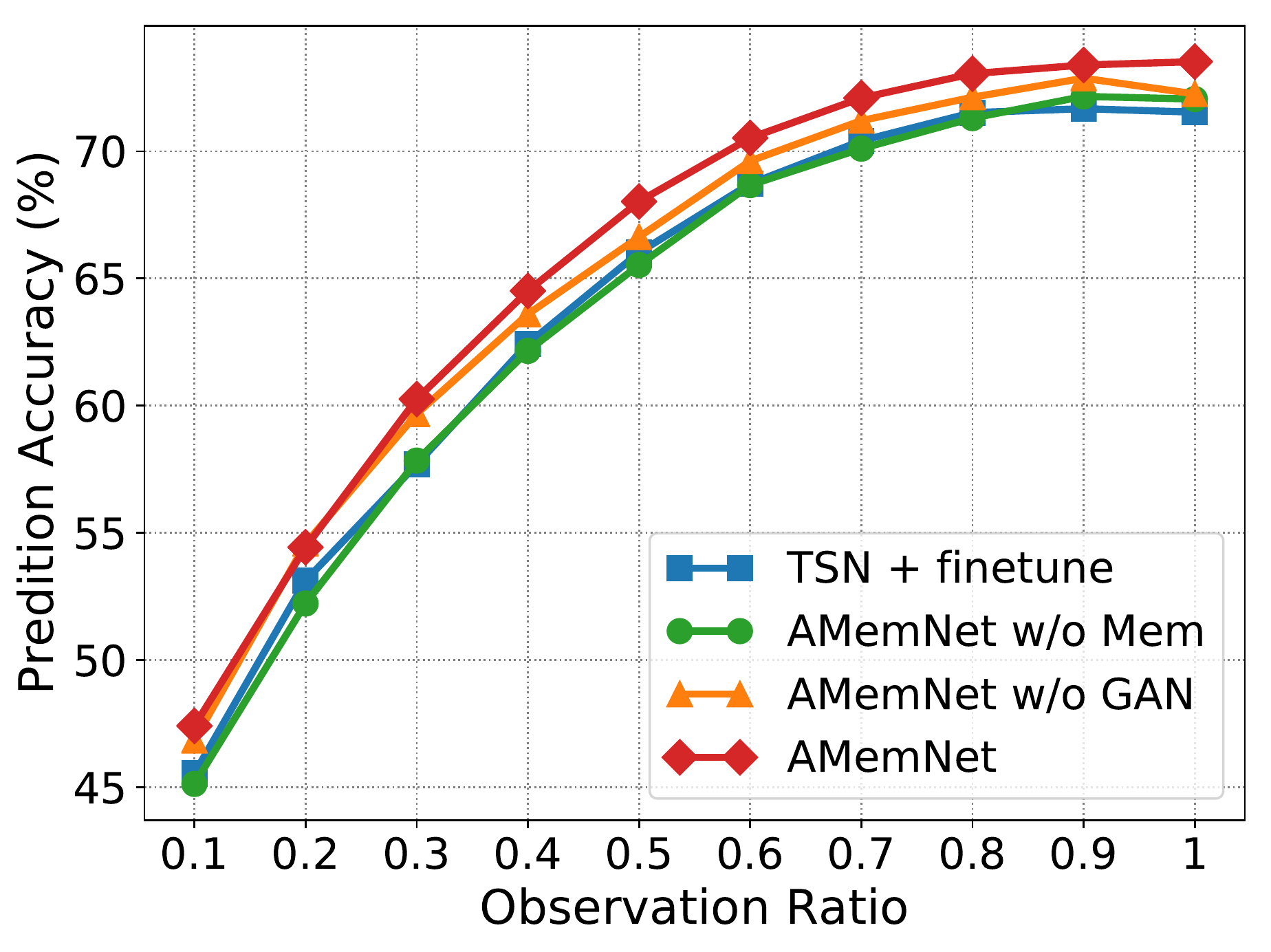}
      \end{minipage}}
     \hfill
     \subfigure[Fusion on HMDB51]{\label{fig:hmdb51-fusion}
      \begin{minipage}[c]{0.32\textwidth}
      \centering
      \includegraphics[width=1\textwidth]{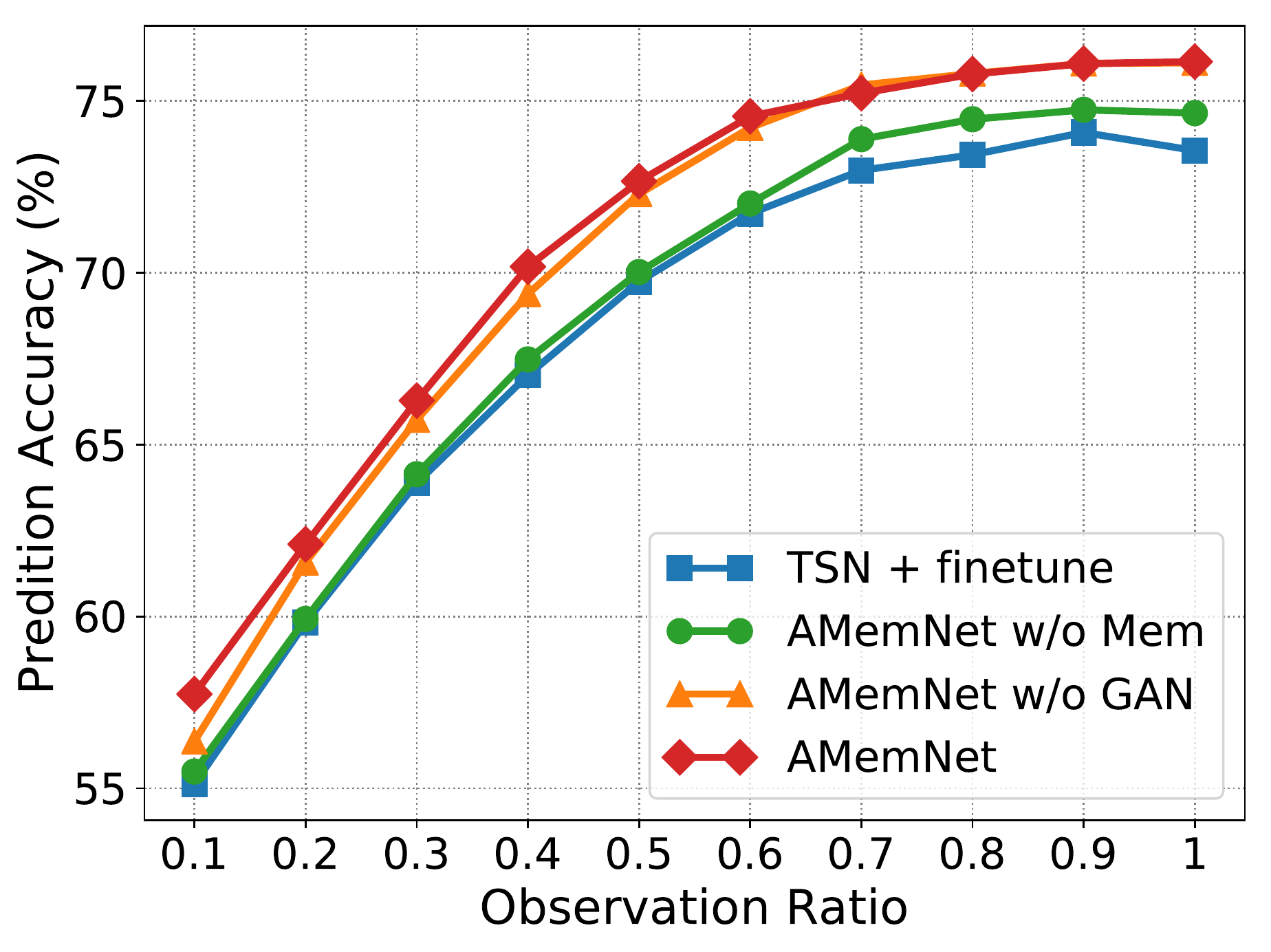}
      \end{minipage}}
    \caption{Ablation study for the proposed AMemNet on the HMDB51 dataset in terms of RGB, Flow and Fusion, respectively}\label{fig:ablation-hmdb51}%\vspace{-0.2cm}
\end{figure*}

We provide 1) more ablation study results (Table~\ref{tab:abl-mth}-\ref{tab:abl-hmdb51} and Fig.~\ref{fig:ablation-hmdb51}) and 2) parameter analyses (Fig.~\ref{fig:para-mem} and Fig.~\ref{fig:para-res}) in the supplementary material. Particularly, we report the averaged results over \emph{three splits} of the UCF101 and HMDB51 datasets for ablation study, respectively, and mainly conduct the parameter analysis on \emph{split 1} of the HMDB51 dataset.

\textbf{Ablation Study}.
The final objective function of the proposed AMemNet model is given by
\begin{small}
\begin{equation*}
\begin{aligned}
\max_{\theta_{D}} &~ \C{L}_{adv} + \lambda_{cls} \C{L}_{cls}^{v},\\
\min_{\theta_{G}} &~ \C{L}_{adv} + \lambda_{cls} \C{L}_{cls}^{x} + \lambda_{rec} \C{L}_{rec},
\end{aligned}
\end{equation*}
\end{small}where $\theta_{G}=\{\theta_{enc}, \theta_{mem}\}$ includes all the trainable parameters for generating $\M{\hat{v}}$ from $\M{x}$, $\theta_{D} = \{\theta_{cls}, \theta_{adv}\}$ parametrizes the discriminator, and $\lambda_{cls}, \lambda_{rec}$ are two trade-off parameters for balancing different terms. 

Table~\ref{tab:abl-mth} shows the overall comparison results between the proposed AMemNet and three ablated models and one baseline. Table~\ref{tab:abl-ucf101} and Table~\ref{tab:abl-hmdb51} provide all the comparison results of each stream on the UCF101 and HMDB51 datasets, respectively. In the proposed AMemNet, the memory-augmented generator $\theta_{G}$ and the adversarial training loss $\C{L}_{adv}$ play the key roles in generating full-video-like features, where $\theta_{G}$ contributes more on the early progress and $\C{L}_{adv}$ leads the late progress. The reconstruction loss $\C{L}_{rec}$ works as a ``regularization'' term for adversarial training, and thus AMemNet w/o Res presents a similar overall performance to our full model. As shown in Table~\ref{tab:abl-hmdb51}, $\C{L}_{rec}$ exhibits a more significant impact on the earlier progress (\emph{e.g.}, $p=0.1$ or $0.2$), since the videos in HMDB51 usually have a lager variance than UCF101. However, overemphasizing $\C{L}_{rec}$ may lower the performance (see $\lambda_{res}=10$ in Fig.~\ref{fig:para-res}).

\begin{figure*}[ht]
     \subfigure[Number of memory slots]{\label{fig:para-mem}
      \begin{minipage}[c]{0.48\textwidth}
      \centering
      \includegraphics[width=0.9\linewidth]{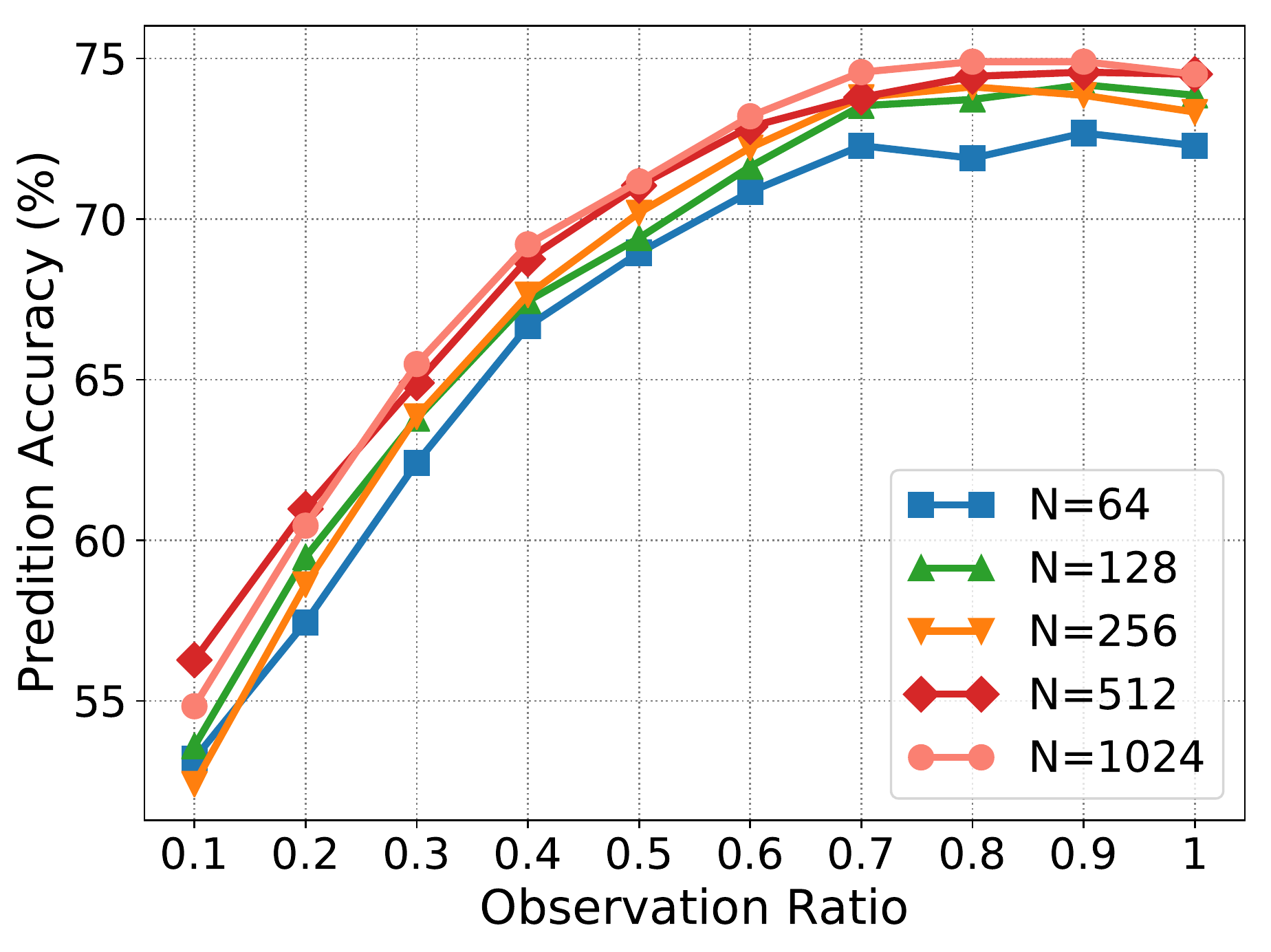}\vspace{-0.2cm}
      \end{minipage}}
     \hfill
     \subfigure[Balancing parameter $\lambda_{res}$]{\label{fig:para-res}
      \begin{minipage}[c]{0.48\textwidth}
      \centering
      \includegraphics[width=0.9\linewidth]{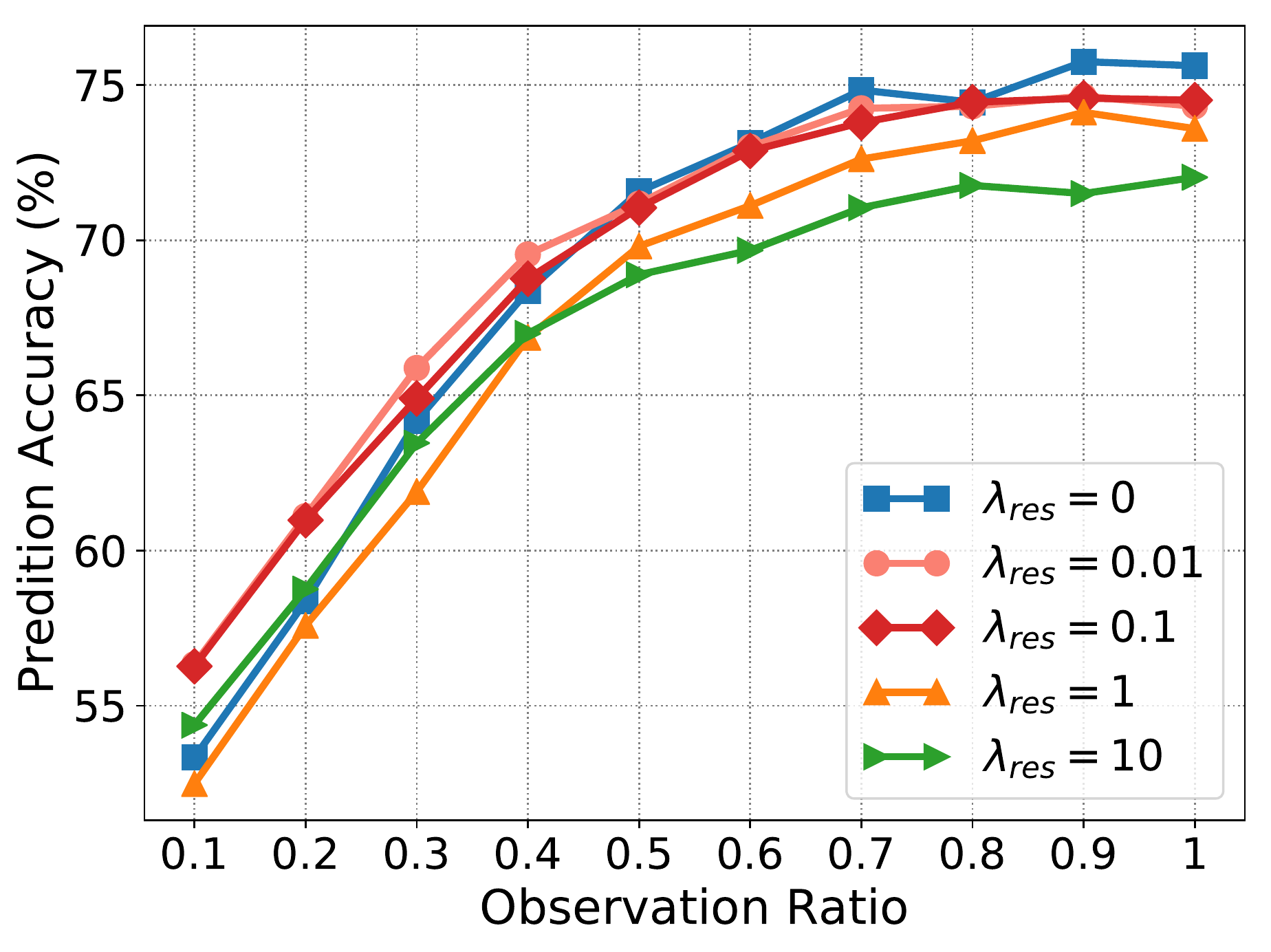}\vspace{-0.2cm}
      \end{minipage}}
    \caption{(a) Parameter analysis on the number of memory slots ($N$). We set $N=$ 64, 128, 256, 512, and 1024, respectively. (b) Parameter analysis on $\lambda_{res}$ for the reconstruction loss $\C{L}_{res}$. We set $\lambda_{res}=$ 0, 0.01, 0.1, 1, and 10, respectively.}
\end{figure*}

\textbf{Parameter Analysis}.
In the experiment, we set the number of memory slots used in $G_{mem}$ as $N=512$ by default. We study the impact of $N$ in Fig.~\ref{fig:para-mem}, suggesting $N \geq 512$ could lead to a stable prediction performance. We fix $\lambda_{cls}=1$ since the classification is the main goal for action prediction, and mainly tune $\lambda_{res}$ in Fig.~\ref{fig:para-res}, which indicates a relatively small $\lambda_{res} \leq 0.1$ is useful for recovering the partial videos. We set $\lambda_{res}=0.1$ as default.

\begin{table*}[ht]
\begin{center}
\caption{Action prediction accuracy (\%) under 10 observation ratios on the UCF101 dataset.}\label{tab:abl-ucf101}%\vspace{-0.2cm}
\scalebox{0.96}{
\begin{tabular}{llcccccccccc}
  \toprule
& Method & $0.1$ & $0.2$ & $0.3$ & $0.4$& $0.5$ & $0.6$ & $0.7$ & $0.8$ & $0.9$& $1.0$\\
\midrule
\multirow{5}{*}{\emph{RGB}}
& TSN + finetune  & $83.81$ & $85.20$ & $86.06$ & $86.81$ & $87.34$ & $88.02$ & $88.40$ & $88.81$ & $88.95$ & $89.36$ \\ 
& AMemNet w/o Mem & $84.92$ & $86.30$ & $87.38$ & $87.80$ & $88.40$ & $88.97$ & $89.47$ & $89.82$ & $90.03$ & $90.04$ \\ 
& AMemNet w/o GAN & $85.33$ & $86.70$ & $87.41$ & $87.98$ & $88.62$ & $88.94$ & $89.58$ & $89.67$ & $89.78$ & $89.86$ \\ 
& AMemNet w/o Res & $86.03$ & $87.27$ & $87.97$ & $88.34$ & $88.90$ & $89.39$ & $89.73$ & $89.93$ & $90.01$ & $90.12$ \\ 
& AMemNet-RGB    & $85.95$ & $87.47$ & $88.21$ & $88.57$ & $89.26$ & $89.51$ & $89.81$ & $89.99$ & $90.06$ & $90.17$ \\
\midrule
\multirow{5}{*}{\emph{Flow}}
& TSN + finetune  & $78.85$ & $84.17$ & $87.14$ & $89.11$ & $90.85$ & $91.94$ & $92.82$ & $93.61$ & $93.70$ & $93.93$ \\ 
& AMemNet w/o Mem & $79.68$ & $84.93$ & $88.15$ & $90.16$ & $91.63$ & $92.80$ & $93.70$ & $94.32$ & $94.65$ & $94.76$ \\
& AMemNet w/o GAN & $81.98$ & $86.87$ & $89.70$ & $90.94$ & $91.91$ & $93.08$ & $93.75$ & $94.20$ & $94.32$ & $94.35$ \\
& AMemNet w/o Res & $83.57$ & $88.00$ & $90.39$ & $91.86$ & $92.79$ & $93.25$ & $93.87$ & $94.34$ & $94.53$ & $94.51$ \\
& AMemNet-Flow    & $83.64$ & $88.32$ & $90.74$ & $92.18$ & $93.20$ & $93.78$ & $94.40$ & $94.75$ & $94.88$ & $94.96$ \\
\midrule
\multirow{5}{*}{\emph{Fusion}}
& TSN + finetune  & $88.88$ & $91.52$ & $93.01$ & $94.05$ & $94.66$ & $95.34$ & $95.64$ & $95.92$ & $95.90$ & $96.00$ \\ 
& AMemNet w/o Mem & $90.17$ & $92.77$ & $94.23$ & $95.03$ & $95.80$ & $96.27$ & $96.52$ & $96.80$ & $96.90$ & $96.76$ \\ 
& AMemNet w/o GAN & $91.47$ & $93.84$ & $94.77$ & $95.45$ & $96.01$ & $96.15$ & $96.55$ & $96.74$ & $96.76$ & $96.72$ \\ 
& AMemNet w/o Res & $92.26$ & $94.15$ & $94.93$ & $95.72$ & $96.16$ & $96.34$ & $96.62$ & $96.63$ & $96.65$ & $96.72$ \\ 
& AMemNet (ours)  & $92.45$ & $94.60$ & $95.55$ & $96.00$ & $96.45$ & $96.67$ & $96.97$ & $96.95$ & $97.07$ & $97.03$ \\
\bottomrule
\end{tabular}}
\end{center}
\end{table*}

\begin{table*}[ht]
\begin{center}
\caption{Action prediction accuracy (\%) under 10 observation ratios on the HMDB51 dataset.}\label{tab:abl-hmdb51}%\vspace{-0.2cm}
\scalebox{0.96}{
\begin{tabular}{llcccccccccc}
  \toprule
& Method & $0.1$ & $0.2$ & $0.3$ & $0.4$& $0.5$ & $0.6$ & $0.7$ & $0.8$ & $0.9$& $1.0$\\
\midrule
\multirow{5}{*}{\emph{RGB}}
& TSN + finetune  & $50.82$ & $54.61$ & $57.03$ & $58.97$ & $60.30$ & $60.98$ & $61.54$ & $61.94$ & $62.23$ & $62.68$ \\ 
& AMemNet w/o Mem & $51.24$ & $54.81$ & $57.63$ & $59.79$ & $61.39$ & $62.67$ & $63.34$ & $63.83$ & $63.92$ & $64.11$ \\ 
& AMemNet w/o GAN & $51.77$ & $55.34$ & $57.98$ & $60.13$ & $61.53$ & $62.56$ & $63.13$ & $63.69$ & $64.28$ & $64.05$ \\ 
& AMemNet w/o Res & $51.51$ & $54.68$ & $57.23$ & $59.63$ & $61.13$ & $61.85$ & $62.32$ & $62.86$ & $63.32$ & $63.34$ \\ 
& AMemNet-RGB     & $52.55$ & $55.52$ & $58.27$ & $60.55$ & $62.53$ & $63.87$ &  $64.41$ & $64.61$ &  $64.99$ & $64.86$ \\
\midrule
\multirow{5}{*}{\emph{Flow}}
& TSN + finetune  & $45.54$ & $53.13$ & $57.69$ & $62.44$ & $66.02$ & $68.72$ & $70.41$ & $71.52$ & $71.68$ & $71.54$ \\ 
& AMemNet w/o Mem & $45.13$ & $52.22$ & $57.84$ & $62.15$ & $65.53$ & $68.67$ & $70.10$ & $71.31$ & $72.16$ & $72.05$ \\ 
& AMemNet w/o GAN & $46.84$ & $54.58$ & $59.68$ & $63.60$ & $66.63$ & $69.62$ & $71.21$ & $72.13$ & $72.88$ & $72.26$ \\ 
& AMemNet w/o Res & $46.65$ & $54.66$ & $59.76$ & $64.33$ & $67.71$ & $70.20$ & $72.02$ & $72.39$ & $72.94$ & $72.57$ \\ 
& AMemNet-Flow    & $47.41$ & $54.43$ & $60.26$ & $64.51$ & $68.03$ & $70.53$ &  $72.10$ & $73.05$ &  $73.39$ & $73.52$ \\
\midrule
\multirow{5}{*}{\emph{Fusion}}
& TSN + finetune  & $55.13$ & $59.82$ & $63.88$ & $67.02$ & $69.74$ & $71.72$ & $72.98$ & $73.43$ & $74.08$ & $73.55$ \\ 
& AMemNet w/o Mem & $55.49$ & $59.93$ & $64.14$ & $67.48$ & $70.02$ & $72.01$ & $73.88$ & $74.46$ & $74.74$ & $74.64$ \\ 
& AMemNet w/o GAN & $56.36$ & $61.57$ & $65.72$ & $69.38$ & $72.29$ & $74.22$ & $75.46$ & $75.79$ & $76.08$ & $76.11$ \\ 
& AMemNet w/o Res & $56.00$ & $60.81$ & $65.78$ & $69.49$ & $72.44$ & $74.58$ & $75.49$ & $75.62$ & $76.73$ & $76.38$ \\ 
& AMemNet (ours)  & $57.74$ & $62.10$ & $66.28$ & $70.17$ & $72.66$ & $74.55$ & $75.22$ & $75.78$ & $76.08$ & $76.14$ \\
\bottomrule
\end{tabular}}
\end{center}
\end{table*}

\end{document}